\documentclass[10pt,twocolumn,letterpaper]{article}

\usepackage[pagenumbers]{cvpr} 

\usepackage{graphicx}
\usepackage{amsmath}
\usepackage{amssymb}
\usepackage{booktabs}
\usepackage{times}
\usepackage{epsfig}
\usepackage{bm}
\usepackage{multirow}
\usepackage{cuted}
\usepackage{capt-of}
\usepackage{caption}
\usepackage{animate}
\usepackage{bbding}
\usepackage{footnote}
\usepackage{xcolor}

\makeatletter
\renewcommand\@makefnmark{\hbox{\@textsuperscript{\normalfont\color{red}\@thefnmark}}}
\makeatother

\usepackage[pagebackref,breaklinks,colorlinks]{hyperref}

\usepackage[capitalize]{cleveref}
\crefname{section}{Sec.}{Secs.}
\Crefname{section}{Section}{Sections}
\Crefname{table}{Table}{Tables}
\crefname{table}{Tab.}{Tabs.}

\graphicspath{{images/}}

\newcommand{\paravspace}{\vspace{-12pt}}

\begin{document}

\title{GRAM: Generative Radiance Manifolds for 3D-Aware Image Generation\vspace{-3pt}}

\author{Yu Deng\thanks{Work done when YD and JX were interns at MSRA.}\,\,$^{1,2}$ \quad Jiaolong Yang$^{2}$ \quad Jianfeng Xiang$^{*3,2}$ \quad Xin Tong$^{2}$ \\
	$^1${Tsinghua University} \quad $^2${Microsoft Research Asia} \quad $^3${USTC} \\
	{\tt\small \{t-yudeng,jiaoyan,v-jxiang,xtong\}@microsoft.com}
}

\graphicspath{{images/}}
\maketitle

\begin{strip}
	\vspace{-55pt}
	\centering
	\includegraphics[width=1\linewidth]{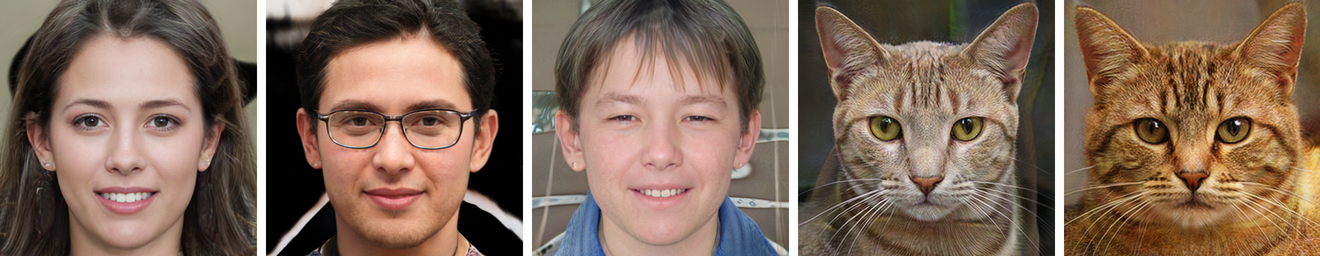}\\
	\animategraphics[loop,autoplay,autopause,width=1\linewidth]{17}{images/teaser/combine_}{0}{39}
	\vspace{-19pt}
	\captionsetup{type=figure,font=small}
	\caption{
		Image samples randomly generated by our method ($256\times 256$ resolution). Trained on unstructured image collections (FFHQ~\cite{karras2019style} and Cats~\cite{zhang2008cat} in this figure), our method can generate view-controllable images that are of high quality (\eg, see the fine details) and strong 3D consistency (\eg, see the correct parallax when view changes). 
		(The second row contains \textbf{\emph{animations}} best viewed in Adobe Reader; more results and code can be found on the \href{https://yudeng.github.io/GRAM/}{\textbf{\emph{project page}}}) 
		\label{fig:teaser}
	}
	\vspace{-3pt}
\end{strip}

\begin{abstract}
	\vspace{-10pt}
	3D-aware image generative modeling aims to generate 3D-consistent images with explicitly controllable camera poses. Recent works have shown promising results by training neural radiance field (NeRF) generators on unstructured 2D images, but still cannot generate highly-realistic images with fine details. A critical reason is that the high memory and computation cost of volumetric representation learning greatly restricts the number of point samples for radiance integration during training. Deficient sampling not only limits the expressive power of the generator to handle fine details but also impedes effective GAN training due to the noise caused by unstable Monte Carlo sampling. We propose a novel approach that regulates point sampling and radiance field learning on 2D manifolds, 
	embodied as a set of learned implicit surfaces in the 3D volume. For each viewing ray, we calculate ray-surface intersections and accumulate their radiance generated by the network. By training and rendering such radiance manifolds, our generator can produce high quality images with realistic fine details and strong visual 3D consistency.
	\href{https://yudeng.github.io/GRAM/}{Code available}.
	\vspace{-18pt}
\end{abstract}

\vspace{-35pt}
\section{Introduction}
\label{sec:intro}
Learning 3D-aware image generation with Generative Adversarial Networks (GAN)~\cite{goodfellow2014generative} has attracted a surge of attention in recent years~\cite{nguyen2019hologan,henzler2019escaping,deng2020disentangled,nguyen2020blockgan,liao2020towards,schwarz2020graf,niemeyer2021giraffe,chan2021pi,devries2021unconstrained}. Given an unstructured 2D image collection, GANs are trained to synthesize geometrically-consistent multiview imagery of novel instances. 
In particular, methods~\cite{henzler2019escaping,schwarz2020graf,chan2021pi} that use the volumetric rendering paradigm~\cite{kajiya1984ray,drebin1988volume} to composite an output image have demonstrated impressive results with more ``strict" 3D consistency 
by virtue of an explicit, physics-based rendering process. 

Notwithstanding the promising results shown by these methods, the image quality still lags far behind traditional 2D image synthesis, for which state-of-the-art GAN models~\cite{karras2019style,karras2020analyzing} can generate high-resolution and photorealistic images. One prominent hurdle is the high computation and memory requirements for training a 
volumetric representation. Methods~\cite{schwarz2020graf,chan2021pi} that use neural radiance field (NeRF)~\cite{mildenhall2020nerf} generators can greatly reduce the complexity of voxel-based approaches~\cite{henzler2019escaping}, but the volume integrations approximated by sampling points along viewing rays are still costly for both training and inference.

This problem becomes even more pronounced in GAN training where a full image (rather than sparse pixels) needs to be rendered to train the discriminator. One workaround is to render patches during training~\cite{schwarz2020graf}, but using a patch discriminator may lead to inferior image generation quality. 
With an image discriminator, the state-of-the art method~\cite{chan2021pi} can only afford training on smaller image resolution and with significantly reduced number of sampling points per ray (typically a few dozens) compared to standard NeRF~\cite{mildenhall2020nerf}. 
However, we observed that radiance integration using Monte Carlo sampling becomes unstable with insufficient samples. The integrated colors among adjacent pixels suffer from intractable noise patterns that are detrimental to GAN training (\eg, see Fig.~\ref{fig:noise}). An even worse issue is that optimizing a full radiance volume requires the sampling to cover both low-frequency regions and high-frequency details, leading to even less sample budget for the latter. Consequently, it is extremely difficult to generate fine details as they simply can be missed by the sampling. 

This paper presents a novel method named Generative Radiance Manifolds (GRAM). 
Different from the previous methods, 
we constrain our point sampling and radiance field learning on 2D manifolds, embodied as a set of implicit surfaces. These implicit surfaces are shared for the trained object category, jointly learned with GAN training, and fixed at inference time. To generate an image, we accumulate the radiance along each ray using ray-surface intersections as point samples. 

There are several advantages of our GRAM method.
First, by confining sampling and radiance learning in a reduced space rather than anywhere in the volume, it greatly facilitates fine detail learning. The network can easily learn to generate thin structures and texture details on the surface manifolds which are guaranteed to have projections on the image and receive supervision during GAN training.
Besides, our generated images are free from the noise pattern caused by inadequate Monte Carlo sampling, as the ray-surface intersections are deterministically calculated and smoothly varying across rays.
Even with very few point samples (\ie, learning very few surfaces), our method can still learn to generate high-quality results.
As a byproduct, at inference time we can render a generated instance in real time by pre-extracting the surfaces with their radiance.

Our implicit surfaces are defined as a set of isosurfaces in a scalar field predicted by a light-weight MLP network. Another MLP for radiance generation is employed, for which we use a structure similar to \cite{chan2021pi}. We extract ray-surface intersections in a differentiable manner, and the whole framework is trained end-to-end using adversarial learning. Orthogonal to our novel radiance manifold design, we also explore network architecture and training method enhancements. In particular, we modify the network structure of \cite{chan2021pi} inspired by \cite{karras2020analyzing} and remove the progressive growing strategy used therein. Progressive growing not only introduces additional hyperparameters to tune but may also lead to degraded image quality shown in traditional 2D GAN~\cite{karras2020analyzing}. We also empirically find that our method generates better results by removing it.

Our method is evaluated on multiple datasets including FFHQ~\cite{karras2019style}, Cats~\cite{zhang2008cat}, and CARLA~\cite{dosovitskiy2017carla,schwarz2020graf}. 
We show that our 3D-aware generation method significantly outperforms the prior art. It can synthesize highly realistic images with geometrically-consistent fine details, which are unseen in previous results. We believe our method makes a significant step towards diminishing the quality gap between 3D-aware generation and traditional 2D image generation.

\section{Related Work}

\paragraph{Neural scene representation and rendering.}
For scene representation and synthesis, a large volume of works~\cite{kulkarni2015deep,dosovitskiy2016learning, tatarchenko2016multi, zhu2017unpaired,isola2017image,eslami2018neural,zhu2018visual, brock2018large,nguyen2018rendernet,park2019semantic,sitzmann2019scene,kim2018deep,sun2019single,bi2019deep,lombardi2019neural,thies2020image} adopt neural networks as a new type of rendering tool due to their ability to synthesize high-quality images without requiring excessive human labor. Among them, earlier works employ convolutional networks for a variety of applications such as novel view synthesis~\cite{hedman2018deep,sitzmann2019deepvoxels,mildenhall2019local,tucker2020single}, image-to-image translation~\cite{bousmalis2017unsupervised,wang2018high,park2019semantic,park2020contrastive}, and controllable image manipulation~\cite{portenier2018faceshop, zhou2021interpreting,bau2019semantic,abdal2021styleflow}.

More recently, plenty of works~\cite{park2019deepsdf,mescheder2019occupancy,sitzmann2020implicit,sitzmann2019scene,mildenhall2020nerf,yariv2020multiview,saito2019pifu,chabra2020deep,niemeyer2020differentiable} leverage implicit neural representations to model 3D scenes using Multi-Layer Perceptrons (MLP). The continuous representation of MLPs brings them the superiority at 3D-level control of image synthesis compared to conventional CNN-based methods. Among these approaches, NeRF~\cite{mildenhall2020nerf,barron2021mip} shows promising results in capturing complex scene structures and synthesizing 3D-consistent images with fine details. Most of the NeRF-based methods~\cite{liu2020neural, park2020deformable, martin2021nerf,peng2021neural,oechsle2021unisurf} 
focus on scene-specific learning tasks where a network is trained to fit a set of posed images of a certain scene. Only a few recent methods~\cite{schwarz2020graf,chan2021pi,niemeyer2021giraffe,hao2021gancraft} work on the image generation task using unconstrained 2D images for supervision. 
This paper proposes a new generative model for improving the image generation quality while maintaining the 3D consistency of generated contents.

\begin{figure*}[t]
	\small
	\centering
	\includegraphics[width=1.0\textwidth]{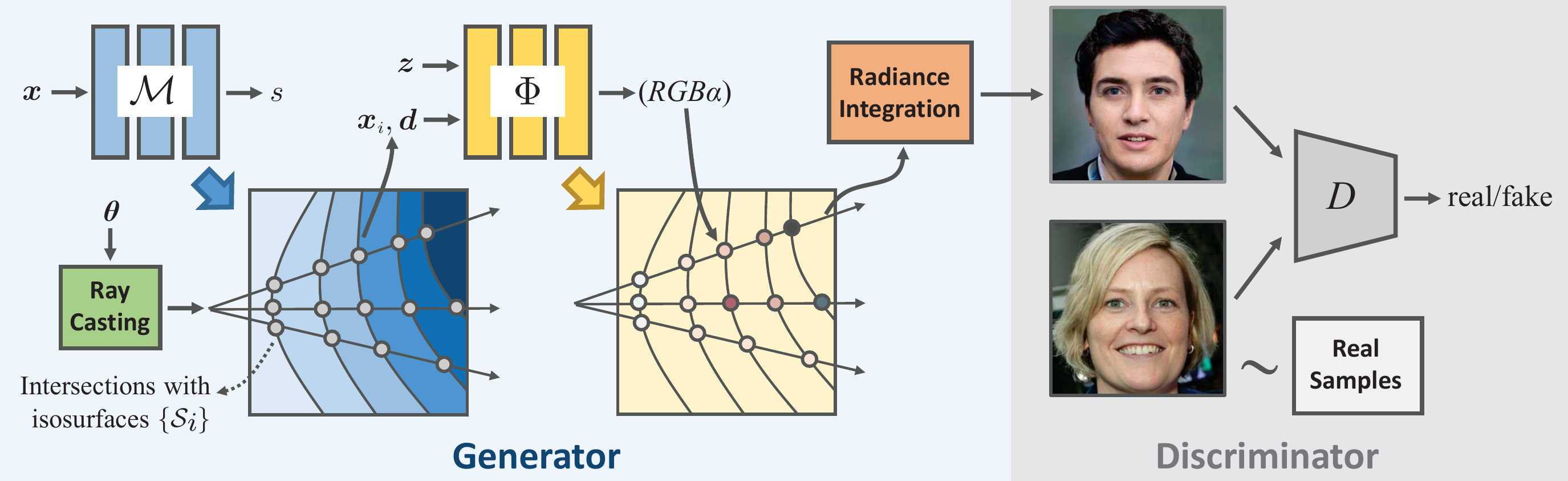}
	\vspace{-18pt}
	\caption{Overview of the GRAM method. The generator $G$ consists of a manifold predictor $\mathcal{M}$ and a radiance generator $\Phi$. $\mathcal{M}$ predicts multiple isosurfaces which define the input domain of $\Phi$. The intersections between camera rays and the isosurfaces are sent to $\Phi$ for color and occupancy prediction. Images are then generated by compositing the color of the points along the ray. 
}
	\label{fig:framework}
	\vspace{-9pt}
\end{figure*}

\vspace{-7pt}
\paragraph{3D-Aware Image Generation.} Given uncontrolled 2D image collections, 3D-aware image generation methods aim to learn a generative model that can explicitly control the camera viewpoint of the generated content. To achieve this goal, the literature mainly follows two directions. The first line of works~\cite{ nguyen2019hologan,liao2020towards,niemeyer2021giraffe,gu2021stylenerf,zhou2021cips} utilize 3D-aware features to represent a scene, and apply a neural renderer, typically a CNN, on top of them for realistic image synthesis. For example, HoloGAN~\cite{nguyen2019hologan} and BlockGAN~\cite{nguyen2020blockgan} learn low-resolution voxel features for objects, project them onto 2D image plane, and apply a StyleGAN-like~\cite{karras2019style} CNN to generate higher-resolution images. Liao~\etal~\cite{liao2020towards} first generate 3D primitives using a 3D generator and then apply a 2D generator with an encoder-decoder structure on the projected features. 
Giraffe~\cite{niemeyer2021giraffe} and GANcraft~\cite{hao2021gancraft} instead use 3D volumetric rendering to generate 2D feature maps for the subsequent image generation. Following a similar idea, some works concurrent to ours~\cite{zhou2021cips,gu2021stylenerf} focus on designing better rendering networks to enable 3D-aware image generation at very high resolution. 
Nevertheless, an inevitable problem of these methods is the sacrifice of exact multi-view consistency due to the learned black-box rendering.

Another group of works~\cite{szabo2019unsupervised,shi2021lifting, schwarz2020graf,chan2021pi,niemeyer2021campari,devries2021unconstrained} seek to learn direct 3D representation of scenes and synthesize images under physical-based rendering process to achieve more strict 3D consistency. \cite{szabo2019unsupervised} and \cite{shi2021lifting} adopt a mesh-based representation and generate images via rasterization. However, they cannot well handle complicated structures with non-Lambertian reflectance such as hair and fur. Recent methods~\cite{schwarz2020graf,chan2021pi,devries2021unconstrained,niemeyer2021campari} use the NeRF representation to synthesize images with high 3D consistency. Still, the expensive computational cost of volumetric representation learning prevents them from generating images with adequate details. 
In this work, we propose a novel approach to learn a generative radiance field on 2D manifolds, and we achieve more realistic image generation with finer details significantly outperforming the previous methods.

\section{Approach}

Given a collection of real images, we learn a 3D-aware image generator $G$ which takes a random noise ${\bm z} \in \mathbb{R}^{d} \sim p_z$ and a camera pose ${\bm \theta} \in \mathbb{R}^{3} \sim p_{\theta}$ as input, and outputs an image $I$ of a synthetic instance under pose ${\bm \theta}$:
\vspace{-2pt}
\begin{equation}
G: ({\bm z},{\bm \theta}) \in \mathbb{R}^{d+3} \rightarrow I \in \mathbb{R}^{H\times W \times 3}.
\vspace{-2pt}
\end{equation}
Figure~\ref{fig:framework} shows the overall structure of $G$, which consists of a manifold predictor $\mathcal{M}$ and a radiance generator $\Phi$. The manifold predictor $\mathcal{M}$ defines a scalar field which derives a reduced domain for radiance generation, which is composed of multiple implicit isosurfaces (Sec.~\ref{sec:mainfold}). Given a latent code ${\bm z}$, the radiance generator $\Phi$ generates the occupancy and color for points on the manifolds (Sec. \ref{sec:radiance}). Images are then generated by integrating the color of the manifold points along each viewing ray (Sec.~\ref{sec:rendering}). The whole method is trained end-to-end 
in an adversarial learning framework (Sec.~\ref{sec:training}). After training, GRAM can render high-quality and 3D-consistent images from different viewpoints.

\subsection{Manifold Predictor}\label{sec:mainfold}

Our manifold predictor $\mathcal{M}$ predicts a reduced space for point sampling and radiance field learning, which is shared across all generated instances. We implement it as a scalar field function which determines a set of isosurfaces.
Specifically,  $\mathcal{M}$ is a light-weight MLP which takes a point $\bm{x}$ as input and predicts a scalar value $s$:
\vspace{-2pt}
\begin{equation}
\mathcal{M}: \bm{x} \in \mathbb{R}^{3} \rightarrow s \in \mathbb{R}. \label{eq:manifold}
\vspace{-2pt}
\end{equation}
Given the predicted scalar field, we obtain $N$ isosurfaces $\{\mathcal{S}_i\}$ with different levels $\{l_i\}$:
\vspace{-2pt}
\begin{equation}
\mathcal{S}_i = \{\bm{x} | \mathcal{M}(\bm{x}) = l_i\}.
\vspace{-2pt}
\end{equation}
These levels are predefined constant values. Note that although the scalar field is defined in the 3D volume of the scene to be rendered, the scalar values per se have no physical meaning and the levels $\{l_i\}$ can be trivially chosen.

We define the input domain of the radiance generator to be on these surfaces. 
Let $\{{\bm x}_i\}$ be the $N$ intersections between a camera ray $\bm{r} = \{\bm{o}+t\bm{d}, t \in [t_n,t_f]\}$ and  $\{\mathcal{S}_i\}$:
\vspace{-5pt}
\begin{equation}
\{{\bm x}_i\} = \{\bm{x}|\bm{x}=\bm{o}+t\bm{d}, \bm{x} \in \{\mathcal{S}_i\}, t \in [t_n,t_f]\}, \label{eq:intersect}
\end{equation}
where $o$ and $d$ are ray origin and direction, and $t_n$ and $t_f$ are the near plane and
far plane parameters. 
We only pass $\{{\bm x}_i\}$ to the radiance generator $\Phi$ for radiance generation and final rendering, as shown in Fig.~\ref{fig:framework}. Since there is no prior knowledge for optimal isosurfaces, we learn them jointly in the generative adversarial training process. 

\begin{figure}[t]
	\small
	\centering
	\includegraphics[width=1.0\columnwidth]{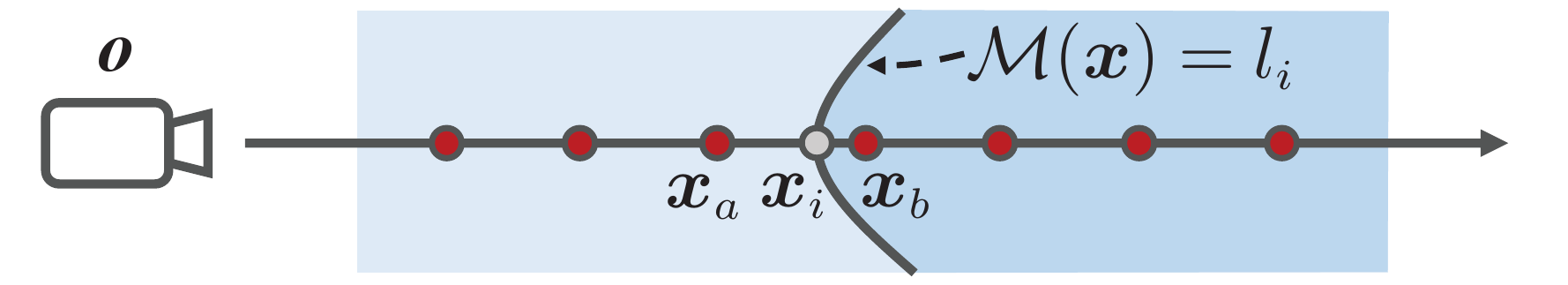}
	\vspace{-19pt}
	\caption{Our differentiable ray-isosurface intersection computation, achieved by linear interpolation between two endpoints of a small interval.}
	\label{fig:intersect}
	\vspace{-6pt}
\end{figure}

Training the manifold predictor $\mathcal{M}$ with GAN necessitates a differentiable scheme for ray-surface intersection computation in order to backpropagate the adversarial loss.
To this end, 
we follow Niemeyer \etal~\cite{niemeyer2020differentiable}'s strategy to calculate the intersections. As shown in Fig.~\ref{fig:intersect}, we evenly sample points along a ray between the near and far planes and feed them to $\mathcal{M}$ to obtain their values $s$. Then we search for the first interval that a certain scalar level $l_i$ falls in, and calculate the intersection using linear interpolation between the two endpoints of the interval via:
\begin{equation}
{\bm x}_i = \frac{l_i-s_a}{s_b-s_a}{\bm x}_b + \frac{s_b-l_i}{s_b-s_a}{\bm x}_a. \label{eq:interpolate}
\end{equation}
We implement $\mathcal{M}$ as a light-weight MLP with 3 hidden layers, and thus dense points 
 (64 points in our implementation) can be sampled to get accurate intersections using Eq.~\eqref{eq:interpolate}.

Random initialization of $\mathcal{M}$ may give rise to highly irregular isosurfaces which is unfavourable for the training process. In this work, we adopt the geometric initialization strategy proposed by Atzmon \etal~\cite{atzmon2020sal} with which the initial isosurfaces are close to spheres.

\subsection{Radiance Generator}\label{sec:radiance}
Given a latent code ${\bm z}$, our radiance generator $\Phi$ generates the radiance for points lying on the learned manifolds. Specifically, $\Phi$ is 
parameterized by an MLP which produces the occupancy $\alpha$ and color ${\bm c} = (R,G,B)$ for a point $\bm{x} \in \mathbb{R}^3$ with view direction $\bm{d}$:
\begin{equation}
\Phi: (\bm{z},\bm{x},\bm{d}) \in \mathbb{R}^{d+6} \rightarrow (\bm{c},\alpha) \in \mathbb{R}^{4}.
\end{equation}
Since radiance is defined on surface manifolds instead of the whole volume in our method, we generate occupancy $\alpha$ instead of volume density $\sigma$ in NeRF, following \cite{oechsle2021unisurf,zhou2018stereo}.

\begin{figure}[t]
	\small
	\centering
	\includegraphics[width=1.0\columnwidth]{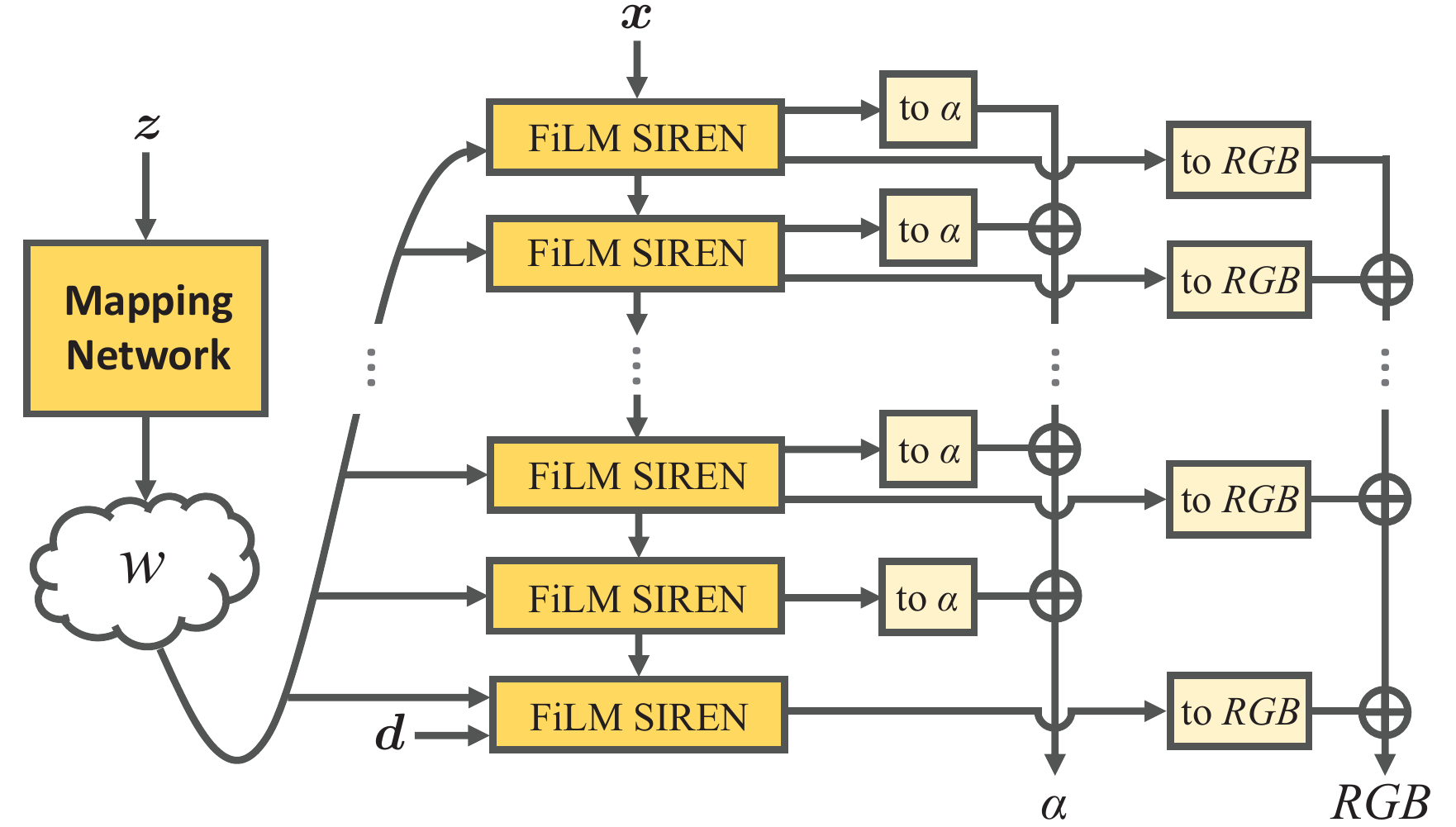}
	\vspace{-15pt}
	\caption{The network structure of radiance generator $\Phi$.}
	\label{fig:backbone}
	\vspace{-7pt}
\end{figure}

The network structure of $\Phi$ is adapted from the FiLM SIREN backbone of \cite{chan2021pi} with some modifications, as presented in Fig.~\ref{fig:backbone}. Inspired by StyleGAN2~\cite{karras2020analyzing}, we use skip connections between output layers at different levels instead of only predicting occupancy and color at the final layer as done in previous methods~\cite{mildenhall2020nerf,chan2021pi}. In this way, different levels of details are now predicted by different output layers and combined together to form the final results. 
This change not only removes the necessity of the progressive growing strategy used in previous methods, but also yields better results in our method as shown in the experiments. 

\begin{figure*}[t]
	\small
	\centering
	\includegraphics[width=1.0\textwidth]{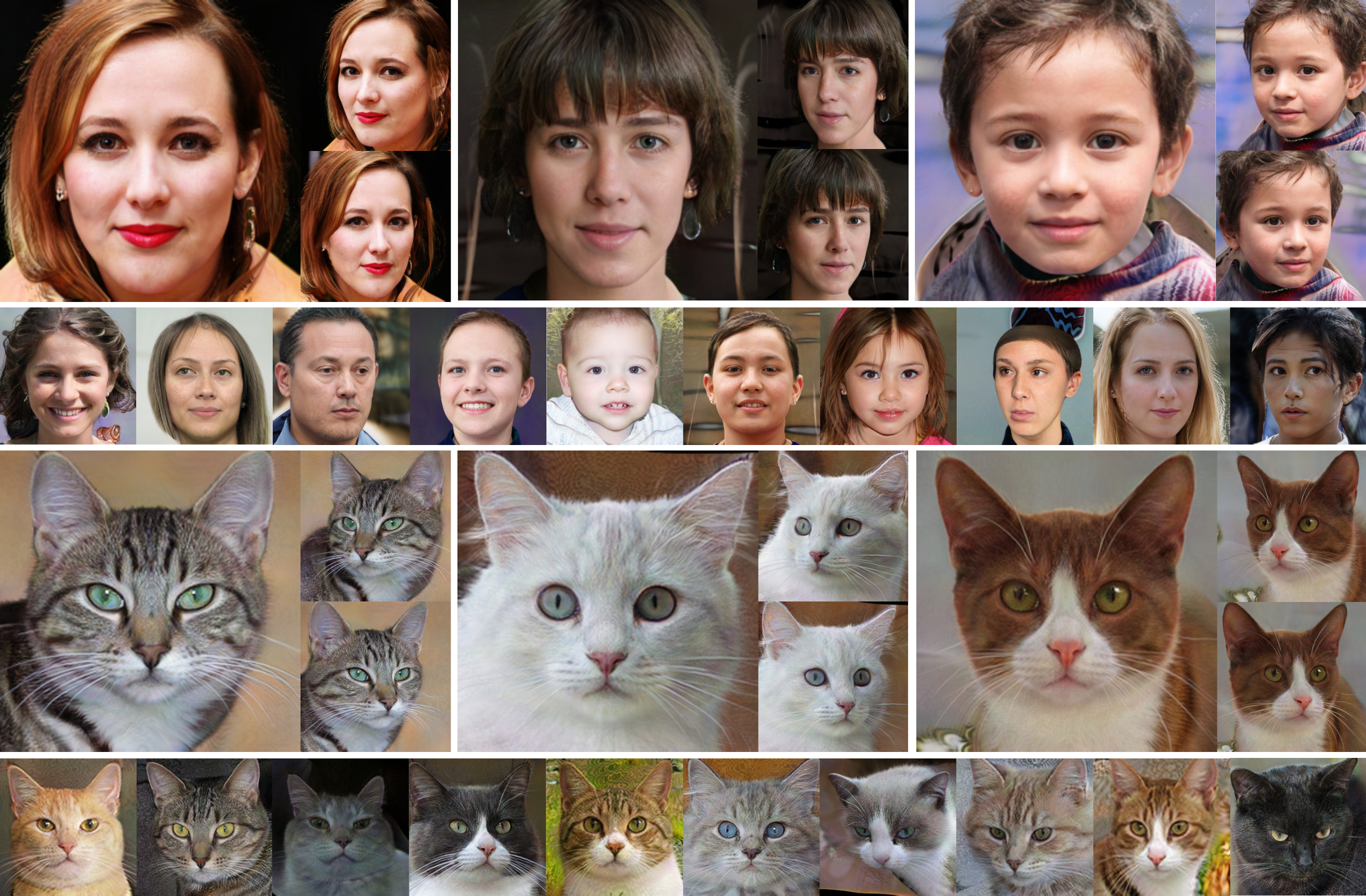}
	\vspace{-18pt}
	\caption{Uncurated $256\times 256$ image samples of human face and cat generated by our method.
	}
	\label{fig:main_results}
	\vspace{-4pt}
\end{figure*}

\subsection{Manifold Rendering}\label{sec:rendering}

For a camera ray $\bm r$ which intersects the surface manifolds at points $\{{\bm x}_i\}$ sorted from near to far following Eq.~\eqref{eq:intersect}, the rendering equation can be written as~\cite{oechsle2021unisurf,zhou2018stereo}:
\vspace{-4pt}
\begin{equation}
\begin{split}
C({\bm r}) = & \sum_{i=1}^{N}T({\bm x}_i)\alpha({\bm x}_i)c({\bm x}_i,{\bm d}) \\
= &  \sum_{i=1}^{N}\prod_{j<i}(1-\alpha({\bm x}_j))\alpha({\bm x}_i)c({\bm x}_i,{\bm d}). \label{eq:render}
\end{split}
\end{equation}
\vspace{-3pt}

Our rendering scheme is clearly different from the original volume rendering in NeRF which applies a hierarchical random sampling strategy (NeRF-H). NeRF-H's sampling points may vary significantly
across adjacent rays due to sampling randomness, resulting in noise patterns on the rendered image (see Fig.~\ref{fig:noise}). By contrast, we only use intersections between camera rays and surface manifolds which are deterministically calculated  and smoothly varying across rays, instead of selecting points in the whole volume space in a Monte Carlo fashion. This helps us eliminate the randomness in image generation and enable training a generator with fewer point samples per ray. Moreover, it greatly facilitates fine detail learning as high-frequency structures and textures can be easily generated on the surface manifolds (see Table~\ref{tab:ablation_sample} and Table~\ref{tab:ablation_number}). 

\subsection{Training Strategy}\label{sec:training}
At training stage, we randomly sample latent code ${\bm z}$ and camera pose ${\bm \theta}$ from prior distributions $p_z$ and $p_\theta$. The generator $G$ synthesizes images with corresponding latent codes and poses as input. We also sample real images from the training data with prior distribution $p_{real}$. As in standard GAN~\cite{goodfellow2014generative}, a discriminator $D$ receives the generated images as well as real images and judge if they are fake or real, for which we use the same CNN structure as in \cite{chan2021pi}. We train all the networks, including the manifold predictor $\mathcal{M}$, the radiance generator $\Phi$ and the discriminator $D$, using non-saturating GAN loss with R1 regularization~\cite{mescheder2018training}:
\\
\vspace{-8pt}
\begin{equation}
\begin{split}
\mathcal{L}(D,G) =& \mathbb{E}_{{\bm z}\sim p_z,{\bm \theta}\sim p_\theta}[f(D(G({\bm z},{\bm \theta})))]\\
& + \mathbb{E}_{I\sim p_{real}}[f(-D(I))+\lambda||\nabla D(I)||^2],
\end{split}
\end{equation}
where $f(u) = \log(1+\exp(u))$ is the Softplus function.

In addition, we find that for certain objects, the training process with only adversarial loss is sometimes sensitive to random initialization. In a few occasions, the learned 3D geometry of convex objects could become concave (see Sec.~\ref{sec:failure}). To tackle this issue, we can optionally add a pose regularization term to enforce the generator to generate images under correct pose:
\begin{equation}
\begin{split}
\mathcal{L}_{pose} =& \mathbb{E}_{{\bm z}\sim p_z,{\bm \theta}\sim p_\theta}||D_p(G({\bm z},{\bm \theta}))-{\bm \theta}||^2 \\
&+ \mathbb{E}_{I\sim p_{real}}||D_p(I)-\hat{{\bm \theta}}||^2, \label{eq:pose}
\end{split}
\end{equation}
where $D_p$ is an additional branch of the discriminator $D$ that predicts the camera pose of a given image, and $\hat{{\bm \theta}}$ is the pose label of a real image. We find that this loss can also slightly improve the image generation quality for objects without the concave geometry issue observed.

\section{Experiments}

\paragraph{Implementation details.} 
We use three datasets for evaluation: FFHQ~\cite{karras2019style},  Cats~\cite{zhang2008cat}, and CARLA~\cite{dosovitskiy2017carla,schwarz2020graf}, which contain 70K high-resolution face images, 10K cat images with various resolutions, and 10K synthetic car images of 16 car models, respectively.
For all experiments, we use the Adam optimizer~\cite{kingma2015adam}, 
and the learning rates are set to $2e{-5}$ for the generator and $2e{-4}$ for the discriminator. 
The models are trained on 8 NVIDIA Tesla V100 GPUs with 32GB memory.
More details can be found in Sec.~\ref{sec:implement}.

\subsection{Generation Results}
\begin{figure*}[t]
	\small
	\centering
	\includegraphics[width=1.0\textwidth]{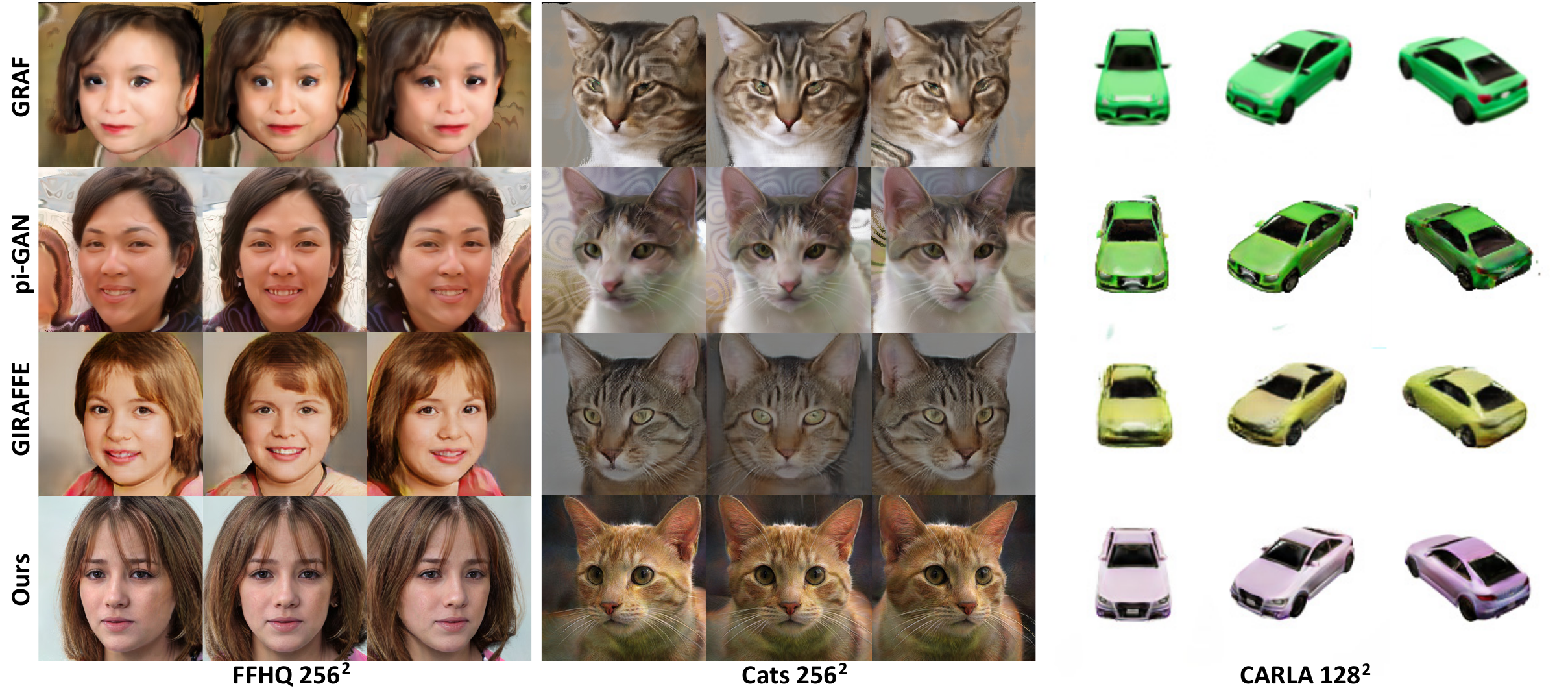}
	\vspace{-19pt}
	\caption{Qualitative comparison with previous 3D-aware image generation methods on three datasets. (\textbf{Best viewed with zoom-in})}
	\label{fig:comparisons}
	\vspace{-9pt}
\end{figure*}

Some random image samples generated by our method are shown in Fig.~\ref{fig:teaser}, \ref{fig:main_results}, and \ref{fig:comparisons}. For face and cat, the model is trained with $256^2$ resolution and $24$ manifold surfaces (\ie, $24$ point samples per ray). For the car images, we train on $128^2$ resolution and use $48$ manifold surfaces. 
As we can see, GRAM is able to generate high-quality images with fine details. Moreover, it allows an explicit control of camera viewpoint and achieves highly consistent results across different views. It even maintains strong visual 3D consistency for very thin structures such as bangs of hair, eyeglass, and whiskers of cat, which show correct parallax corresponding to realistic 3D geometry. Note that \emph{3D consistency is best viewed with  animations}, 
which can be found on our \href{https://yudeng.github.io/GRAM/}{\emph{project page}}.

\begin{figure}[t]
	\small
	\centering
	\includegraphics[width=0.95\columnwidth]{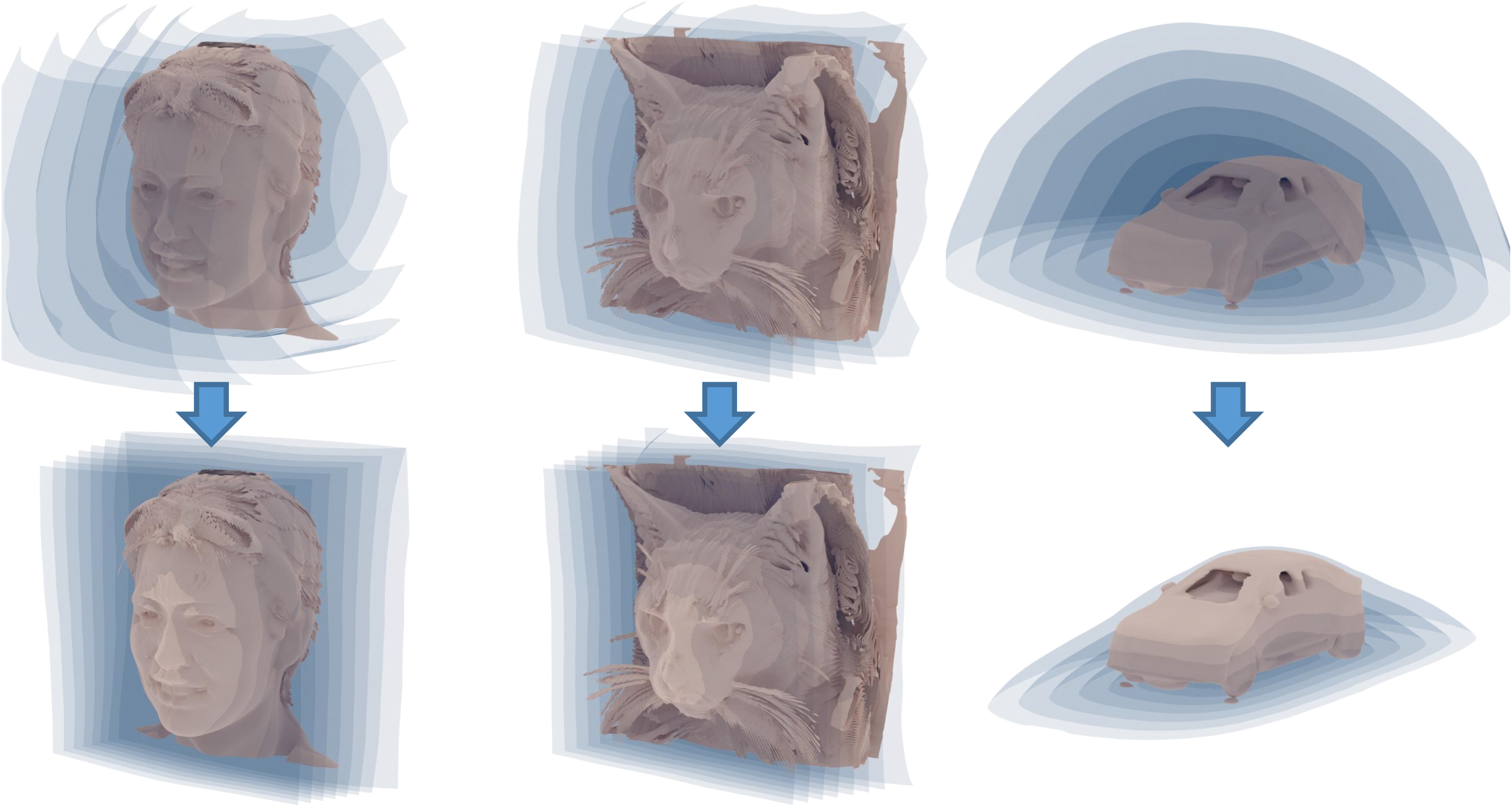}
	\vspace{-10pt}
	\caption{Initial (top) and final (bottom) surface manifolds learned on three datasets. Eight evenly-sampled surfaces are visualized here. To show the relative position of the surfaces in the 3D object space, we also visualize an extracted 3D shape for reference.}
	\label{fig:manifold_vis}
	\vspace{-2pt}
\end{figure}

\begin{figure}[t]
	\small
	\centering
	\includegraphics[width=1.0\columnwidth]{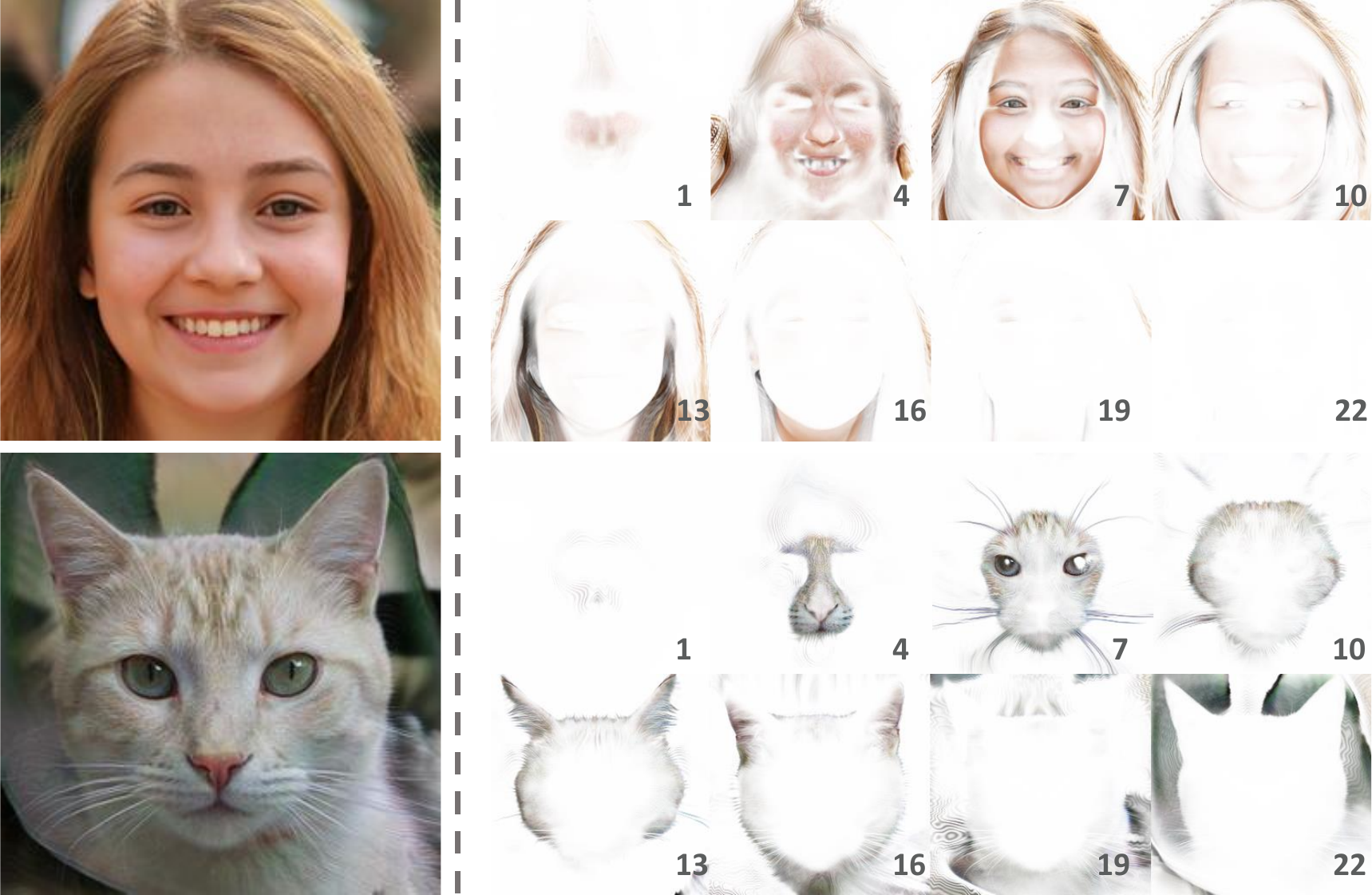}
	\vspace{-19pt}
	\caption{Visualization of generated radiance on the surface manifolds. Eight evenly sampled surfaces from front to back are shown.}
	\label{fig:slice}
	\vspace{-8pt}
\end{figure}

\paravspace
\paragraph{Visualization of surface manifolds.} Figure~\ref{fig:manifold_vis} shows the learned surface manifolds on the three datasets. Initially, the surfaces have near-spherical shapes and are positioned across the whole volume. 
After training, the surfaces for face and cat are tightened and exhibit small curvatures. The surfaces for car are also tightened but maintain a curving structure that covers the car geometry. The face and cat images from FFHQ~\cite{karras2019style} and Cats~\cite{zhang2008cat} only have small angle variations; most of them are nearly frontal. In this case, near-planar surfaces are enough to render a generated instance. In contrast, the camera viewpoints of the car images from CARLA~\cite{schwarz2020graf} are uniformly distributed on the upper hemisphere (\ie, $360^\circ$ azimuth and $90^\circ$ elevation angles). Such a wide viewpoint range necessities curved surfaces to ensure good rendering results from different views.

Figure~\ref{fig:slice} shows the radiance predicted on the manifolds with two examples.
We evenly sample surfaces from front to back and render the color patterns on them with their contribution to the final image as opacity. As  shown in the figures, the network is able to learn high-frequency details and thin structures (\eg whiskers) on the manifolds.

\begin{figure}[t]
	\small
	\centering
	\includegraphics[width=1.0\columnwidth]{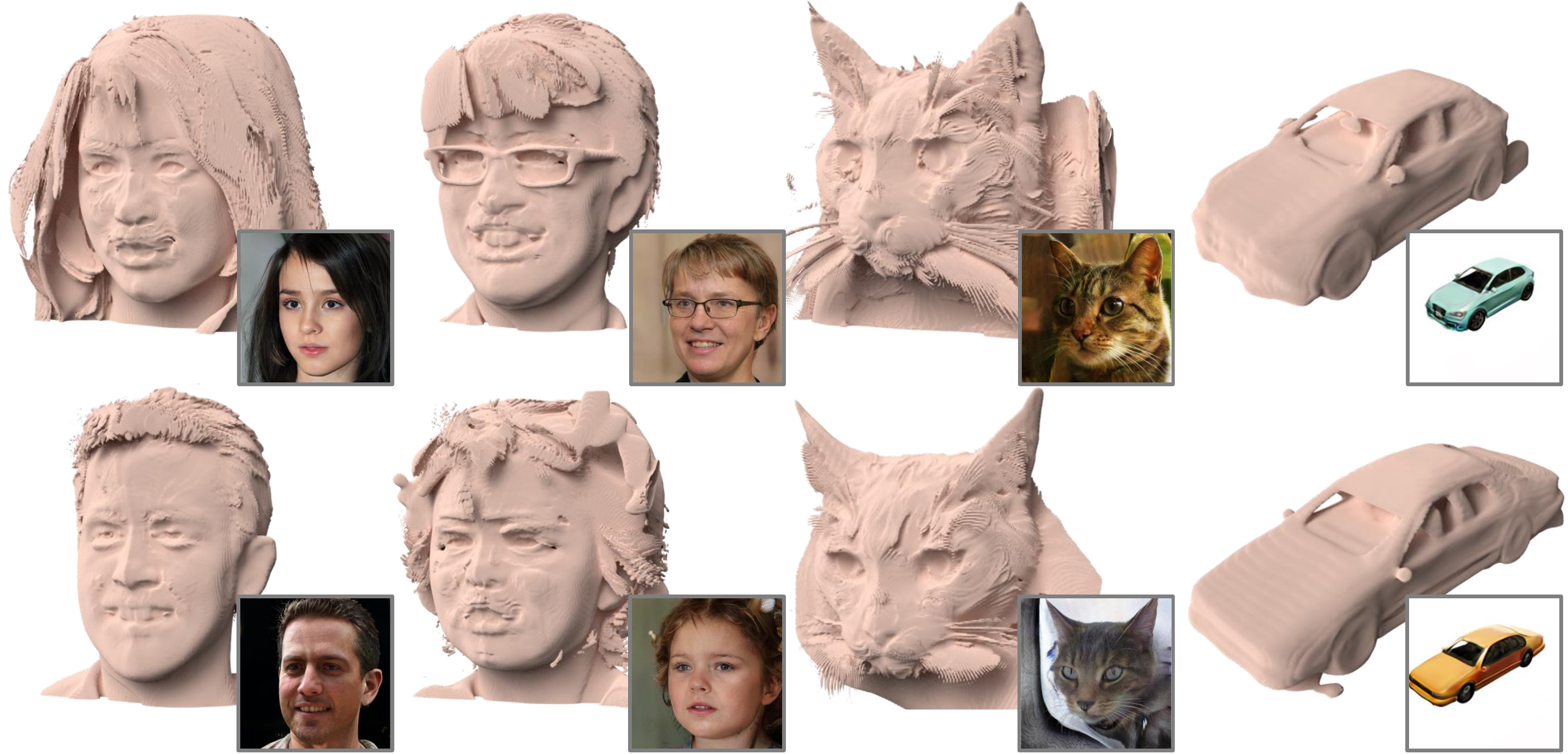}
	\vspace{-18pt}
	\caption{Extracted proxy 3D shapes of the generated instances.}
	\label{fig:shapes}
	\vspace{-9pt}
\end{figure}

\begin{figure}[t]
	\small
	\centering
	\includegraphics[width=1.0\columnwidth]{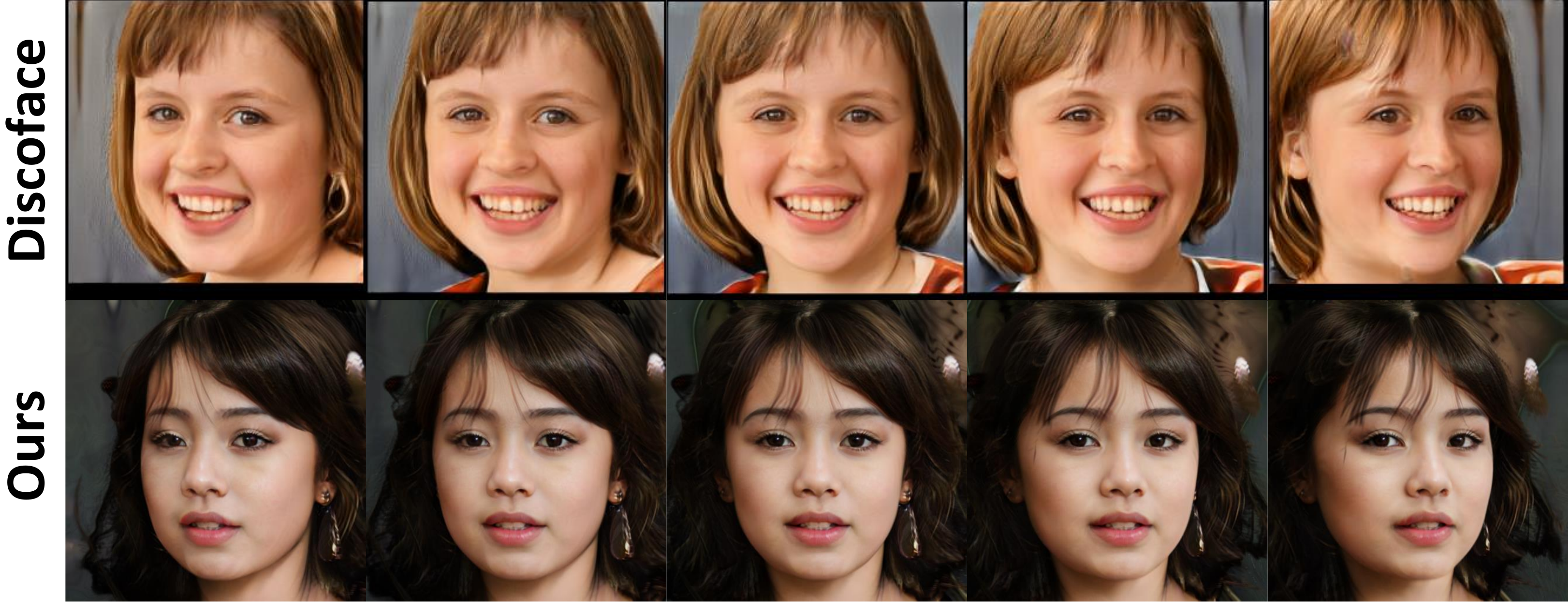}
	\vspace{-19pt}
	\caption{Qualitative comparison with a controllable face image generation method DiscofaceGAN. (\textbf{Best viewed with zoom-in})}
	\label{fig:comparison_disco}
	\vspace{-6pt}
\end{figure}

\paravspace
\paragraph{Visualization of 3D geometry.}
Although our method confines the input domain of the radiance field on 2D manifolds, we can still extract proxy 3D shapes of the generated objects using the volume-based marching cubes algorithm~\cite{lorensen1987marching}. 
Figure~\ref{fig:shapes} shows the proxy 3D shapes of several generated instances. It can be observed that our method produces high-quality geometry with detailed structures well depicted, which is the key to achieve strong visual 3D consistency across different views for not only low-frequency regions but also fine details.

\subsection{Comparison with Previous Methods}
We compare GRAM with three state-of-the-art 3D-aware image generation approaches: GRAF~\cite{schwarz2020graf}, pi-GAN~\cite{chan2021pi}, and GIRAFFE~\cite{niemeyer2021giraffe}. Experiments are conducted using the official implementation provided by the authors. For GRAF and GIRAFFE, we modify the camera pose distribution according to different datasets, and leave other configurations unchanged. For pi-GAN, we follow the authors' settings that use 24, 48, and 96 sampling points for FFHQ, Cats, and CARLA respectively, for both training and testing. Note that for our method, we use 24 surfaces for FFHQ and Cats, and 48 surfaces for CARLA.

We further compare GRAM with a face-specific controllable image generation approach: DiscofaceGAN~\cite{deng2020disentangled}, which uses a 2D CNN as the generator and achieves pose control with the guidance of a prior 3D face model~\cite{paysan20093d}.

\vspace{-1pt}
\paravspace
\paragraph{Qualitative comparison.} Figure~\ref{fig:comparisons} shows the visual comparison between GRAM and other methods. As we can see, GRAF and pi-GAN struggle to generate high-frequency details such as the texture of hair and fur. GIRAFFE produces images with finer details, but it suffers from 3D inconsistency (\eg, see hair region of the woman) due to the use of a CNN renderer . Our method achieves the best visual quality with realistic details and remarkable 3D consistency. See Fig.~\ref{fig:compare_more} and our \href{https://yudeng.github.io/GRAM/}{\emph{project page}} for more results.

Figure~\ref{fig:comparison_disco} shows the qualitative comparison between GRAM and DiscofaceGAN. While DiscofaceGAN can generate realistic face images and explicitly control their camera poses, it cannot well maintain the 3D consistency (\eg, see the bangs). By contrast, GRAM achieves strong 3D consistency under comparable generation quality without requiring extra 3D face priors.

\paravspace
\paragraph{Quantitative comparison.} We evaluate the image quality using the Fr\'echet Inception Distances (FID)~\cite{heusel2017gans} and Kernel Inception Distances (KID)~\cite{binkowski2018demystifying} between $20$K randomly generated images and $20$K sampled real images. 
Table~\ref{tab:comparisons} shows that 
we significantly improve the two metrics compared to GRAF and pi-GAN, which also use NeRF generators. We even achieve lower FID and KID compare to GIRAFFE which applies a refinement CNN after the NeRF rendering to achieve better image quality. GIRAFFE is trained on a single GPU following its original implementation.

\subsection{Ablation Study}
We further conduct ablation study to validate the efficacy of our method designs. For efficiency, all experiments are conducted on FFHQ with $128^2$ resolution. Unless otherwise specified, we use $24$ points per ray for these experiments.

\paravspace
\paragraph{Sampling methods.} We compare our manifold sampling strategy with several baseline methods as shown in Table~\ref{tab:ablation_sample}. \textit{NeRF-H} is the original hierarchical sampling strategy used in NeRF~\cite{mildenhall2020nerf} and pi-GAN~\cite{chan2021pi}. \textit{Planes} denotes using intersections between camera rays and multiple parallel planes placed across the volume. \textit{Spherical (init)} denotes sphere-like surfaces obtained from the geometric initialization \cite{atzmon2020sal} and fixed during training.
Compare to the alternatives, our learnable manifolds yield the best image quality in terms of FID metrics. \textit{NeRF-H} has a large performance gap with the others, indicating its deficiency under limited sample points.
Our method outperforms \textit{Planes} and \textit{Spherical (init)}, which demonstrates the advantage of using learnable surfaces that can better fit the trained object category.

\begin{table}[t]
    \centering
    \caption{Quantitative comparisons on three datasets using FID and KID$\times100$ between 20K generated images and 20K real images. Results of StyleGAN2~\cite{karras2020analyzing} are included for reference. $\dag$: Evaluated using pre-trained models provided by the authors.}       \label{tab:comparisons}
    \vspace{-7pt}
    \begin{tabular}{c|cc|cc|cc}
    \toprule[1pt]
    & \multicolumn{2}{c|}{\!FFHQ 256$^2$\!} & \multicolumn{2}{c|}{\!Cats 256$^2$\!} & \multicolumn{2}{c}{\!\!CARLA 128$^2$\!\!}\\
    Methods & \!FID\! & \!KID\! & \!FID\! & \!KID\! & \!FID\! & \!KID\!\\
    \hline
    \!StyleGAN2\! & 6.97 & 0.17 & 8.41 & 0.32 & 10.4 & 0.47\\
    \hline
    GRAF & 73.0 & 5.89 & 59.5 & 4.59 & \!\!32.1$^\dag$\!\!\!\!&\!\!1.84$^\dag$\!\!\!\!\\
    pi-GAN & 55.2 & 4.13 & \!\!53.7\!\! & \!\!4.35\!\! & \!\!36.0$^\dag$\!\!\!\! & \!\!2.08$^\dag$\!\!\!\!\\
    GIRAFFE & \!\!32.6$^\dag$\!\!\!\! & \!\!2.24$^\dag$\!\!\!\! & 20.7 & 1.14 & \!\!105\textcolor{red}{$^\text{1}$}\!\!\!\! & \!\!7.19\!\!\\
    Ours & \textbf{17.9} & \textbf{0.84} & \textbf{14.6} & \textbf{0.75} & \textbf{26.3}& \textbf{1.15}\\
     & \!(\textbf{14.5}\textcolor{red}{$^\text{2}$})\!\!\! & \!(\textbf{0.65}\textcolor{red}{$^\text{2}$})\!\!\! &  &  & & \\
    \bottomrule[1pt]
    \end{tabular}

\vspace{-8pt}
\end{table}

\addtocounter{footnote}{-2}
\footnotetext[1]{We tried our best to train GIRAFFE on CARLA using multiple different settings and report the best result we obtained.}
\addtocounter{footnote}{-1}
\footnotetext[2]{Updated results as of May 2022, obtained with larger batchsize and more iterations for training (batchsize 16 with 120K iterations $\rightarrow$ batchsize 32 with 150K iterations)}

\begin{figure}[t]
	\small
	\centering
	\includegraphics[width=1.0\columnwidth]{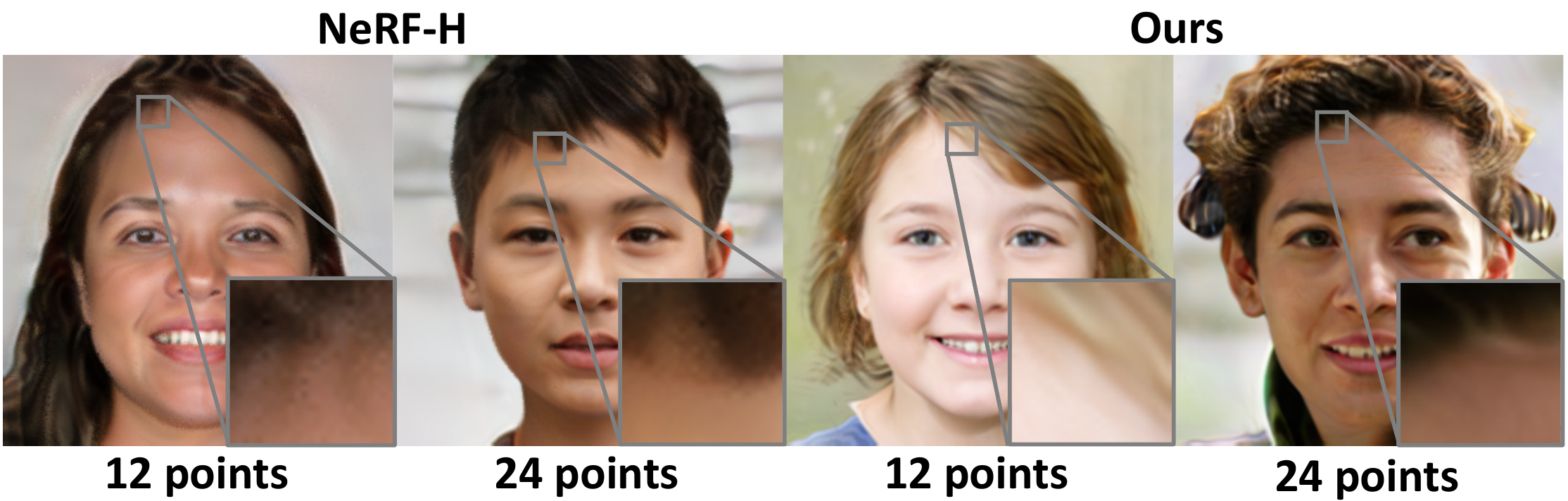}
	\vspace{-14pt}
	\caption{Images generated using NeRF-H~\cite{mildenhall2020nerf,chan2021pi} sampling contain noise patterns under limited point samples whereas ours are noise-free. \textbf{(Best viewed with zoom-in)}}
	\label{fig:noise}
	\vspace{-4pt}
\end{figure}

\paravspace
\paragraph{Number of surface manifolds.} We further evaluate the generation quality of GRAM when training with different number of surfaces. For a reference, we also train models using the hierarchical sampling strategy \textit{NeRF-H} with same number of sampling points for each ray. Table~\ref{tab:ablation_number} shows that our method can generate high quality results using as few as 6 surfaces, and adding more gradually improves the quality. In contrast, training with \textit{NeRF-H} largely fails with less than 12 points as indicated by the high FIDs, due to the difficulty to handle high-frequency details as well as the noise brought by inadequate sampling (Fig.~\ref{fig:noise}).
Even using 48 points, its generation quality is still worse than ours with 6 surfaces. In addition, it tends to learn unreasonable geometry with concave human foreheads, which rarely happens in our case (see Fig.~\ref{fig:number} for visual results).

\paravspace
\paragraph{Influence of pose regularization.} Table~\ref{tab:ablation_pose} shows the effect of using pose labels of real images in Eq.~\eqref{eq:pose} during training. For human face, our method produces slightly better results using the real pose regularization. In contrast, the hierarchical sampling strategy is unstable without real pose as guidance, leading to much worse results.

\begin{table}[t]
	\centering
	\small
	\caption{Ablation study on different point sampling strategies ($24$ points used for each ray; $12$ coarse and $12$ fine points for NeRF-H)}      \label{tab:ablation_sample}
	\vspace{-7pt}
	\begin{tabular}{c|c|c|c|c}
		\toprule[1pt]
		& \!\!NeRF-H~\!\!\cite{mildenhall2020nerf,chan2021pi}\!\! & Planes & \!\!\! Spherical (init) \!\!\! & Ours\\
		\hline
		FID 5K & 35.4 & 28.3 & 27.8 & \textbf{25.8}\\
		\bottomrule[1pt]
	\end{tabular}
	\vspace{0pt}
\end{table}

\begin{table}[t]
	\centering
	\small
	\caption{Ablation study on number of sampling points per ray.}      \label{tab:ablation_number}
	\vspace{-7pt}
	\begin{tabular}{c|c|c|c|c|c|c}
		\toprule[1pt]
		\multicolumn{2}{c|}{\!Number of points\!} & 6 & 12 & 24 & 36 & 48 \\
		\hline
		\multirow{2}{*}{\!\!FID 5K\!\!}& \!\!NeRF-H~\!\!\cite{mildenhall2020nerf,chan2021pi}\!\! & 117 & 62.6 & 35.4 & 32.9 & 30.0\\
		& Ours & \textbf{27.4} & \textbf{27.0} & \textbf{25.8} & \textbf{25.8} & \textbf{25.2}\\
		\bottomrule[1pt]
	\end{tabular}
	\vspace{0pt}
\end{table}

\begin{table}[t]
\centering
\small
\caption{Ablation study on pose regularization.}      \label{tab:ablation_pose}
\vspace{-7pt}
\begin{tabular}{c|c|c|c}
	\toprule[1pt]
& \!Real pose\!& \!\!NeRF-H~\!\! \cite{mildenhall2020nerf,chan2021pi}\!\! & Ours  \\
	\hline
	\multirow{2}{*}{FID 5K} & \XSolidBrush & 44.4 & 26.4  \\
	& \Checkmark & 35.4 & 25.8  \\
	\bottomrule[1pt]
\end{tabular}
\vspace{0pt}
\end{table}

\begin{table}[t]
	\centering
	\small
	\caption{Ablation study on training strategy and network structure.}      \label{tab:ablation_train}
	\vspace{-7pt}
	\begin{tabular}{c|c|c|c}
		\toprule[1pt]
		& Base & - PG & + Skip (Ours)\\
		\hline
		FID 5K & 30.6 & 28.8 & \textbf{25.8} \\
		\bottomrule[1pt]
	\end{tabular}
	\vspace{0pt}
\end{table}

\paravspace
\paragraph{Training strategy and network structure.} As shown in Table~\ref{tab:ablation_train}, we first train our GRAM model with the network structure proposed in~\cite{chan2021pi} and the progressive growing strategy from $32^2$ resolution following \cite{chan2021pi}, which is the \textit{Base} setting. Then we switch to the non-progressive growing strategy by training a model from scratch using $128^2$ resolution. Finally, we add skip connections in the network structure as depicted in Fig.~\ref{fig:backbone}. The improvements on FID clearly demonstrate the advantages of our design.

\subsection{Real-time multiview synthesis.}
For objects generated by GRAM, we can achieve real-time free-view  rendering thanks to our radiance manifold design. 
Spefically, we pre-extract the surface manifolds using marching cubes~\cite{lorensen1987marching} and store the radiance on them.
With an efficient mesh rasterizer~\cite{laine2020modular}, we achieve 180FPS free-view rendering of $256^2$ images on a Nvidia Tesla V100 GPU.

\section{Conclusions}
We presented a novel approach for 3D-aware image generation. The core idea is to regulate point sampling and radiance learning on 2D manifolds for the radiance generator. Extensive experiments have shown its superiority over previous methods on both generation quality and 3D consistency. We believe our method takes a large step towards generating 3D-aware virtual contents for real applications.

\vspace{5pt}
\paravspace
\paragraph{Ethics and responsible AI considerations.} 
The goal of this paper is to study generative modelling of the 3D objects from 2D images, and to provide a method for generating multi-view images of non-existing, virtual objects. It is not intended to manipulate existing images nor to create content that is used to mislead or deceive. This method does not have understanding and control of the generated content. Thus, adding targeted facial expressions or mouth movements is out of the scope of this work. However, the method, like all other related AI image generation techniques, could still potentially be misused for impersonating humans. Currently, the images generated by this method contain visual artifacts, unnatural texture patterns, and other unpredictable failures that can be spotted by humans and fake image detection algorithms. We also plan to investigate applying this technology for advancing 3D- and video-based forgery detection.  

\vspace{5pt}
\paravspace
\paragraph{Limitations and future works.} 
Under constrained sampling budgets, our shared surfaces across the whole class can cause certain artifacts (see Sec.~\ref{sec:failure}) and limit our method to object categories sharing similar geometry. It may not well handle complex 3D scenes of multiple subjects with diverse structures. Learning instance-specific manifolds is a possible solution in the future. Besides, the generation quality and speed of GRAM still falls behind traditional 2D GANs. Better representations could be explored to further improve the fidelity and efficiency.

\vspace{5pt}
\paravspace
\paragraph{Acknowledgements.} We thank Harry Shum for the fruitful advice and discussion to improve the paper.  

{\small
\bibliographystyle{ieee_fullname}
\bibliography{egbib}

\begin{thebibliography}{10}\itemsep=-1pt

\bibitem{abdal2021styleflow}
Rameen Abdal, Peihao Zhu, Niloy~J Mitra, and Peter Wonka.
\newblock Styleflow: Attribute-conditioned exploration of stylegan-generated
  images using conditional continuous normalizing flows.
\newblock {\em ACM Transactions on Graphics}, 40(3):1--21, 2021.

\bibitem{atzmon2020sal}
Matan Atzmon and Yaron Lipman.
\newblock Sal: Sign agnostic learning of shapes from raw data.
\newblock In {\em IEEE/CVF Conference on Computer Vision and Pattern
  Recognition}, pages 2565--2574, 2020.

\bibitem{barron2021mip}
Jonathan~T Barron, Ben Mildenhall, Matthew Tancik, Peter Hedman, Ricardo
  Martin-Brualla, and Pratul~P Srinivasan.
\newblock Mip-nerf: A multiscale representation for anti-aliasing neural
  radiance fields.
\newblock In {\em IEEE/CVF International Conference on Computer Vision}, 2021.

\bibitem{bau2019semantic}
David Bau, Hendrik Strobelt, William Peebles, Jonas Wulff, Bolei Zhou, Jun-Yan
  Zhu, and Antonio Torralba.
\newblock Semantic photo manipulation with a generative image prior.
\newblock {\em ACM Transactions on Graphics}, 38(4):1--11, 2019.

\bibitem{bi2019deep}
Sai Bi, Kalyan Sunkavalli, Federico Perazzi, Eli Shechtman, Vladimir~G Kim, and
  Ravi Ramamoorthi.
\newblock Deep cg2real: Synthetic-to-real translation via image
  disentanglement.
\newblock In {\em International Conference on Computer Vision}, pages
  2730--2739, 2019.

\bibitem{binkowski2018demystifying}
Miko{\l}aj Bi{\'n}kowski, Danica~J Sutherland, Michael Arbel, and Arthur
  Gretton.
\newblock Demystifying mmd gans.
\newblock In {\em International Conference on Learning Representations}, 2018.

\bibitem{bousmalis2017unsupervised}
Konstantinos Bousmalis, Nathan Silberman, David Dohan, Dumitru Erhan, and Dilip
  Krishnan.
\newblock Unsupervised pixel-level domain adaptation with generative
  adversarial networks.
\newblock In {\em IEEE Conference on Computer Vision and Pattern Recognition},
  pages 3722--3731, 2017.

\bibitem{brock2018large}
Andrew Brock, Jeff Donahue, and Karen Simonyan.
\newblock Large scale gan training for high fidelity natural image synthesis.
\newblock In {\em International Conference on Learning Representations}, 2019.

\bibitem{bulat2017far}
Adrian Bulat and Georgios Tzimiropoulos.
\newblock How far are we from solving the 2d \& 3d face alignment problem?(and
  a dataset of 230,000 3d facial landmarks).
\newblock In {\em IEEE International Conference on Computer Vision}, pages
  1021--1030, 2017.

\bibitem{chabra2020deep}
Rohan Chabra, Jan~E Lenssen, Eddy Ilg, Tanner Schmidt, Julian Straub, Steven
  Lovegrove, and Richard Newcombe.
\newblock Deep local shapes: Learning local sdf priors for detailed 3d
  reconstruction.
\newblock In {\em European Conference on Computer Vision}, pages 608--625,
  2020.

\bibitem{chan2021pi}
Eric~R Chan, Marco Monteiro, Petr Kellnhofer, Jiajun Wu, and Gordon Wetzstein.
\newblock pi-gan: Periodic implicit generative adversarial networks for
  3d-aware image synthesis.
\newblock In {\em IEEE/CVF Conference on Computer Vision and Pattern
  Recognition}, pages 5799--5809, 2021.

\bibitem{deng2019arcface}
Jiankang Deng, Jia Guo, Niannan Xue, and Stefanos Zafeiriou.
\newblock Arcface: Additive angular margin loss for deep face recognition.
\newblock In {\em IEEE/CVF Conference on Computer Vision and Pattern
  Recognition}, pages 4690--4699, 2019.

\bibitem{deng2020disentangled}
Yu Deng, Jiaolong Yang, Dong Chen, Fang Wen, and Xin Tong.
\newblock Disentangled and controllable face image generation via 3{D}
  imitative-contrastive learning.
\newblock In {\em IEEE/CVF Conference on Computer Vision and Pattern
  Recognition}, pages 5154--5163, 2020.

\bibitem{deng2019accurate}
Yu Deng, Jiaolong Yang, Sicheng Xu, Dong Chen, Yunde Jia, and Xin Tong.
\newblock Accurate 3d face reconstruction with weakly-supervised learning: From
  single image to image set.
\newblock In {\em IEEE/CVF Conference on Computer Vision and Pattern
  Recognition Workshops}, pages 0--0, 2019.

\bibitem{devries2021unconstrained}
Terrance DeVries, Miguel~Angel Bautista, Nitish Srivastava, Graham~W Taylor,
  and Joshua~M Susskind.
\newblock Unconstrained scene generation with locally conditioned radiance
  fields.
\newblock In {\em IEEE/CVF International Conference on Computer Vision}, 2021.

\bibitem{dosovitskiy2017carla}
Alexey Dosovitskiy, German Ros, Felipe Codevilla, Antonio Lopez, and Vladlen
  Koltun.
\newblock Carla: An open urban driving simulator.
\newblock In {\em Conference on Robot Learning}, pages 1--16, 2017.

\bibitem{dosovitskiy2016learning}
Alexey Dosovitskiy, Jost~Tobias Springenberg, Maxim Tatarchenko, and Thomas
  Brox.
\newblock Learning to generate chairs, tables and cars with convolutional
  networks.
\newblock {\em IEEE Transactions on Pattern Analysis and Machine Intelligence},
  39(4):692--705, 2016.

\bibitem{drebin1988volume}
Robert~A Drebin, Loren Carpenter, and Pat Hanrahan.
\newblock Volume rendering.
\newblock {\em ACM SIGGRAPH}, 22(4):65--74, 1988.

\bibitem{eslami2018neural}
SM~Ali Eslami, Danilo~Jimenez Rezende, Frederic Besse, Fabio Viola, Ari~S
  Morcos, Marta Garnelo, Avraham Ruderman, Andrei~A Rusu, Ivo Danihelka, Karol
  Gregor, et~al.
\newblock Neural scene representation and rendering.
\newblock {\em Science}, 360(6394):1204--1210, 2018.

\bibitem{goodfellow2014generative}
Ian Goodfellow, Jean Pouget-Abadie, Mehdi Mirza, Bing Xu, David Warde-Farley,
  Sherjil Ozair, Aaron Courville, and Yoshua Bengio.
\newblock Generative adversarial nets.
\newblock {\em Advances in Neural Information Processing Systems}, 27, 2014.

\bibitem{gu2021stylenerf}
Jiatao Gu, Lingjie Liu, Peng Wang, and Christian Theobalt.
\newblock Stylenerf: A style-based 3d-aware generator for high-resolution image
  synthesis.
\newblock {\em arXiv preprint arXiv:2110.08985}, 2021.

\bibitem{hao2021gancraft}
Zekun Hao, Arun Mallya, Serge Belongie, and Ming-Yu Liu.
\newblock Gancraft: Unsupervised 3d neural rendering of minecraft worlds.
\newblock In {\em IEEE/CVF International Conference on Computer Vision}, 2021.

\bibitem{hedman2018deep}
Peter Hedman, Julien Philip, True Price, Jan-Michael Frahm, George Drettakis,
  and Gabriel Brostow.
\newblock Deep blending for free-viewpoint image-based rendering.
\newblock {\em ACM Transactions on Graphics}, 37(6):1--15, 2018.

\bibitem{henzler2019escaping}
Philipp Henzler, Niloy~J Mitra, and Tobias Ritschel.
\newblock Escaping plato's cave: 3d shape from adversarial rendering.
\newblock In {\em IEEE/CVF International Conference on Computer Vision}, pages
  9984--9993, 2019.

\bibitem{heusel2017gans}
Martin Heusel, Hubert Ramsauer, Thomas Unterthiner, Bernhard Nessler, and Sepp
  Hochreiter.
\newblock Gans trained by a two time-scale update rule converge to a local nash
  equilibrium.
\newblock In {\em Advances in Neural Information Processing Systems}, pages
  6626--6637, 2017.

\bibitem{isola2017image}
Phillip Isola, Jun-Yan Zhu, Tinghui Zhou, and Alexei~A Efros.
\newblock Image-to-image translation with conditional adversarial networks.
\newblock In {\em IEEE Conference on Computer Vision and Pattern Recognition},
  pages 1125--1134, 2017.

\bibitem{kajiya1984ray}
James~T Kajiya and Brian~P Von~Herzen.
\newblock Ray tracing volume densities.
\newblock {\em ACM SIGGRAPH}, 18(3):165--174, 1984.

\bibitem{karras2019style}
Tero Karras, Samuli Laine, and Timo Aila.
\newblock A style-based generator architecture for generative adversarial
  networks.
\newblock In {\em IEEE/CVF Conference on Computer Vision and Pattern
  Recognition}, pages 4401--4410, 2019.

\bibitem{karras2020analyzing}
Tero Karras, Samuli Laine, Miika Aittala, Janne Hellsten, Jaakko Lehtinen, and
  Timo Aila.
\newblock Analyzing and improving the image quality of stylegan.
\newblock In {\em IEEE/CVF Conference on Computer Vision and Pattern
  Recognition}, pages 8110--8119, 2020.

\bibitem{kim2018deep}
Hyeongwoo Kim, Pablo Garrido, Ayush Tewari, Weipeng Xu, Justus Thies, Matthias
  Niessner, Patrick P{\'e}rez, Christian Richardt, Michael Zollh{\"o}fer, and
  Christian Theobalt.
\newblock Deep video portraits.
\newblock {\em ACM Transactions on Graphics}, 37(4):1--14, 2018.

\bibitem{kingma2015adam}
Diederik~P Kingma and Jimmy Ba.
\newblock Adam: A method for stochastic optimization.
\newblock In {\em International Conference on Learning Representations}, 2015.

\bibitem{kulkarni2015deep}
Tejas~D Kulkarni, Will Whitney, Pushmeet Kohli, and Joshua~B Tenenbaum.
\newblock Deep convolutional inverse graphics network.
\newblock In {\em Advances in Neural Information Processing Systems}, 2015.

\bibitem{laine2020modular}
Samuli Laine, Janne Hellsten, Tero Karras, Yeongho Seol, Jaakko Lehtinen, and
  Timo Aila.
\newblock Modular primitives for high-performance differentiable rendering.
\newblock {\em ACM Transactions on Graphics}, 39(6):1--14, 2020.

\bibitem{liao2020towards}
Yiyi Liao, Katja Schwarz, Lars Mescheder, and Andreas Geiger.
\newblock Towards unsupervised learning of generative models for 3d
  controllable image synthesis.
\newblock In {\em IEEE/CVF Conference on Computer Vision and Pattern
  Recognition}, pages 5871--5880, 2020.

\bibitem{liu2020neural}
Lingjie Liu, Jiatao Gu, Kyaw~Zaw Lin, Tat-Seng Chua, and Christian Theobalt.
\newblock Neural sparse voxel fields.
\newblock In {\em Advances in Neural Information Processing Systems}, 2020.

\bibitem{lombardi2019neural}
Stephen Lombardi, Tomas Simon, Jason Saragih, Gabriel Schwartz, Andreas
  Lehrmann, and Yaser Sheikh.
\newblock Neural volumes: learning dynamic renderable volumes from images.
\newblock {\em ACM Transactions on Graphics}, 38(4):1--14, 2019.

\bibitem{lorensen1987marching}
William~E Lorensen and Harvey~E Cline.
\newblock Marching cubes: A high resolution 3d surface construction algorithm.
\newblock {\em ACM SIGGRAPH}, 21(4):163--169, 1987.

\bibitem{martin2021nerf}
Ricardo Martin-Brualla, Noha Radwan, Mehdi~SM Sajjadi, Jonathan~T Barron,
  Alexey Dosovitskiy, and Daniel Duckworth.
\newblock Nerf in the wild: Neural radiance fields for unconstrained photo
  collections.
\newblock In {\em IEEE/CVF Conference on Computer Vision and Pattern
  Recognition}, pages 7210--7219, 2021.

\bibitem{mescheder2018training}
Lars Mescheder, Andreas Geiger, and Sebastian Nowozin.
\newblock Which training methods for gans do actually converge?
\newblock In {\em International Conference on Machine Learning}, pages
  3481--3490, 2018.

\bibitem{mescheder2019occupancy}
Lars Mescheder, Michael Oechsle, Michael Niemeyer, Sebastian Nowozin, and
  Andreas Geiger.
\newblock Occupancy networks: Learning 3d reconstruction in function space.
\newblock In {\em IEEE/CVF Conference on Computer Vision and Pattern
  Recognition}, pages 4460--4470, 2019.

\bibitem{mildenhall2019local}
Ben Mildenhall, Pratul~P Srinivasan, Rodrigo Ortiz-Cayon, Nima~Khademi
  Kalantari, Ravi Ramamoorthi, Ren Ng, and Abhishek Kar.
\newblock Local light field fusion: Practical view synthesis with prescriptive
  sampling guidelines.
\newblock {\em ACM Transactions on Graphics}, 38(4):1--14, 2019.

\bibitem{mildenhall2020nerf}
Ben Mildenhall, Pratul~P Srinivasan, Matthew Tancik, Jonathan~T Barron, Ravi
  Ramamoorthi, and Ren Ng.
\newblock Nerf: Representing scenes as neural radiance fields for view
  synthesis.
\newblock In {\em European Conference on Computer Vision}, pages 405--421.
  Springer, 2020.

\bibitem{nguyen2018rendernet}
Thu Nguyen-Phuoc, Chuan Li, Stephen Balaban, and Yong-Liang Yang.
\newblock Rendernet: A deep convolutional network for differentiable rendering
  from 3d shapes.
\newblock In {\em Advances in Neural Information Processing Systems}, 2018.

\bibitem{nguyen2019hologan}
Thu Nguyen-Phuoc, Chuan Li, Lucas Theis, Christian Richardt, and Yong-Liang
  Yang.
\newblock Hologan: Unsupervised learning of 3{D} representations from natural
  images.
\newblock In {\em IEEE/CVF International Conference on Computer Vision}, pages
  7588--7597, 2019.

\bibitem{nguyen2020blockgan}
Thu Nguyen-Phuoc, Christian Richardt, Long Mai, Yong-Liang Yang, and Niloy
  Mitra.
\newblock {BlockGAN}: Learning 3d object-aware scene representations from
  unlabelled images.
\newblock In {\em Advances in Neural Information Processing Systems}, 2020.

\bibitem{niemeyer2021campari}
Michael Niemeyer and Andreas Geiger.
\newblock Campari: Camera-aware decomposed generative neural radiance fields.
\newblock In {\em International Conference on 3D Vision}, pages 951--961. IEEE,
  2021.

\bibitem{niemeyer2021giraffe}
Michael Niemeyer and Andreas Geiger.
\newblock Giraffe: Representing scenes as compositional generative neural
  feature fields.
\newblock In {\em IEEE/CVF Conference on Computer Vision and Pattern
  Recognition}, pages 11453--11464, 2021.

\bibitem{niemeyer2020differentiable}
Michael Niemeyer, Lars Mescheder, Michael Oechsle, and Andreas Geiger.
\newblock Differentiable volumetric rendering: Learning implicit 3d
  representations without 3d supervision.
\newblock In {\em IEEE/CVF Conference on Computer Vision and Pattern
  Recognition}, pages 3504--3515, 2020.

\bibitem{oechsle2021unisurf}
Michael Oechsle, Songyou Peng, and Andreas Geiger.
\newblock Unisurf: Unifying neural implicit surfaces and radiance fields for
  multi-view reconstruction.
\newblock In {\em IEEE/CVF International Conference on Computer Vision}, 2021.

\bibitem{park2019deepsdf}
Jeong~Joon Park, Peter Florence, Julian Straub, Richard Newcombe, and Steven
  Lovegrove.
\newblock Deepsdf: Learning continuous signed distance functions for shape
  representation.
\newblock In {\em IEEE/CVF Conference on Computer Vision and Pattern
  Recognition}, pages 165--174, 2019.

\bibitem{park2020deformable}
Keunhong Park, Utkarsh Sinha, Jonathan~T Barron, Sofien Bouaziz, Dan~B Goldman,
  Steven~M Seitz, and Ricardo Martin-Brualla.
\newblock Deformable neural radiance fields.
\newblock In {\em IEEE/CVF International Conference on Computer Vision}, 2021.

\bibitem{park2020contrastive}
Taesung Park, Alexei~A Efros, Richard Zhang, and Jun-Yan Zhu.
\newblock Contrastive learning for unpaired image-to-image translation.
\newblock In {\em European Conference on Computer Vision}, pages 319--345.
  Springer, 2020.

\bibitem{park2019semantic}
Taesung Park, Ming-Yu Liu, Ting-Chun Wang, and Jun-Yan Zhu.
\newblock Semantic image synthesis with spatially-adaptive normalization.
\newblock In {\em IEEE/CVF Conference on Computer Vision and Pattern
  Recognition}, pages 2337--2346, 2019.

\bibitem{paysan20093d}
Pascal Paysan, Reinhard Knothe, Brian Amberg, Sami Romdhani, and Thomas Vetter.
\newblock A 3{D} face model for pose and illumination invariant face
  recognition.
\newblock In {\em IEEE International Conference on Advanced Video and Signal
  Based Surveillance}, pages 296--301, 2009.

\bibitem{peng2021neural}
Sida Peng, Yuanqing Zhang, Yinghao Xu, Qianqian Wang, Qing Shuai, Hujun Bao,
  and Xiaowei Zhou.
\newblock Neural body: Implicit neural representations with structured latent
  codes for novel view synthesis of dynamic humans.
\newblock In {\em IEEE/CVF Conference on Computer Vision and Pattern
  Recognition}, pages 9054--9063, 2021.

\bibitem{portenier2018faceshop}
Tiziano Portenier, Qiyang Hu, Attila Szabo, Siavash Arjomand, Paolo Favaro, and
  Matthias Zwicker.
\newblock Faceshop: Deep sketch-based image editing.
\newblock {\em ACM Transactions on Graphics}, 37(4):1--13, 2018.

\bibitem{saito2019pifu}
Shunsuke Saito, Zeng Huang, Ryota Natsume, Shigeo Morishima, Angjoo Kanazawa,
  and Hao Li.
\newblock Pifu: Pixel-aligned implicit function for high-resolution clothed
  human digitization.
\newblock In {\em IEEE/CVF International Conference on Computer Vision}, pages
  2304--2314, 2019.

\bibitem{schwarz2020graf}
Katja Schwarz, Yiyi Liao, Michael Niemeyer, and Andreas Geiger.
\newblock Graf: Generative radiance fields for 3d-aware image synthesis.
\newblock In {\em Advances in Neural Information Processing Systems}, 2020.

\bibitem{shi2021lifting}
Yichun Shi, Divyansh Aggarwal, and Anil~K Jain.
\newblock Lifting 2d stylegan for 3d-aware face generation.
\newblock In {\em IEEE/CVF Conference on Computer Vision and Pattern
  Recognition}, pages 6258--6266, 2021.

\bibitem{sitzmann2020implicit}
Vincent Sitzmann, Julien Martel, Alexander Bergman, David Lindell, and Gordon
  Wetzstein.
\newblock Implicit neural representations with periodic activation functions.
\newblock {\em Advances in Neural Information Processing Systems}, 33, 2020.

\bibitem{sitzmann2019deepvoxels}
Vincent Sitzmann, Justus Thies, Felix Heide, Matthias Nie{\ss}ner, Gordon
  Wetzstein, and Michael Zollhofer.
\newblock Deepvoxels: Learning persistent 3d feature embeddings.
\newblock In {\em IEEE/CVF Conference on Computer Vision and Pattern
  Recognition}, pages 2437--2446, 2019.

\bibitem{sitzmann2019scene}
Vincent Sitzmann, Michael Zollh{\"o}fer, and Gordon Wetzstein.
\newblock Scene representation networks: continuous 3d-structure-aware neural
  scene representations.
\newblock In {\em Advances in Neural Information Processing Systems}, pages
  1121--1132, 2019.

\bibitem{sun2019single}
Tiancheng Sun, Jonathan~T Barron, Yun-Ta Tsai, Zexiang Xu, Xueming Yu, Graham
  Fyffe, Christoph Rhemann, Jay Busch, Paul~E Debevec, and Ravi Ramamoorthi.
\newblock Single image portrait relighting.
\newblock {\em ACM Transactions on Graphics}, 38(4):79--1, 2019.

\bibitem{szabo2019unsupervised}
Attila Szab{\'o}, Givi Meishvili, and Paolo Favaro.
\newblock Unsupervised generative 3d shape learning from natural images.
\newblock {\em arXiv preprint arXiv:1910.00287}, 2019.

\bibitem{tatarchenko2016multi}
Maxim Tatarchenko, Alexey Dosovitskiy, and Thomas Brox.
\newblock Multi-view 3d models from single images with a convolutional network.
\newblock In {\em European Conference on Computer Vision}, pages 322--337,
  2016.

\bibitem{thies2020image}
Justus Thies, Michael Zollh{\"o}fer, Christian Theobalt, Marc Stamminger, and
  Matthias Nie{\ss}ner.
\newblock Image-guided neural object rendering.
\newblock In {\em International Conference on Learning Representations}, 2020.

\bibitem{tucker2020single}
Richard Tucker and Noah Snavely.
\newblock Single-view view synthesis with multiplane images.
\newblock In {\em IEEE/CVF Conference on Computer Vision and Pattern
  Recognition}, pages 551--560, 2020.

\bibitem{wang2018high}
Ting-Chun Wang, Ming-Yu Liu, Jun-Yan Zhu, Andrew Tao, Jan Kautz, and Bryan
  Catanzaro.
\newblock High-resolution image synthesis and semantic manipulation with
  conditional gans.
\newblock In {\em IEEE Conference on Computer Vision and Pattern Recognition},
  pages 8798--8807, 2018.

\bibitem{yariv2020multiview}
Lior Yariv, Yoni Kasten, Dror Moran, Meirav Galun, Matan Atzmon, Ronen Basri,
  and Yaron Lipman.
\newblock Multiview neural surface reconstruction by disentangling geometry and
  appearance.
\newblock In {\em Advances in Neural Information Processing Systems}, 2020.

\bibitem{zhang2018unreasonable}
Richard Zhang, Phillip Isola, Alexei~A Efros, Eli Shechtman, and Oliver Wang.
\newblock The unreasonable effectiveness of deep features as a perceptual
  metric.
\newblock In {\em IEEE/CVF Conference on Computer Vision and Pattern
  Recognition}, pages 586--595, 2018.

\bibitem{zhang2008cat}
Weiwei Zhang, Jian Sun, and Xiaoou Tang.
\newblock Cat head detection-how to effectively exploit shape and texture
  features.
\newblock In {\em European Conference on Computer Vision}, pages 802--816,
  2008.

\bibitem{zhou2021interpreting}
Bolei Zhou.
\newblock Interpreting generative adversarial networks for interactive image
  generation.
\newblock {\em arXiv preprint arXiv:2108.04896}, 2021.

\bibitem{zhou2021cips}
Peng Zhou, Lingxi Xie, Bingbing Ni, and Qi Tian.
\newblock Cips-3d: A 3d-aware generator of gans based on
  conditionally-independent pixel synthesis.
\newblock {\em arXiv preprint arXiv:2110.09788}, 2021.

\bibitem{zhou2018stereo}
Tinghui Zhou, Richard Tucker, John Flynn, Graham Fyffe, and Noah Snavely.
\newblock Stereo magnification: learning view synthesis using multiplane
  images.
\newblock {\em ACM Transactions on Graphics}, 37(4):1--12, 2018.

\bibitem{zhu2017unpaired}
Jun-Yan Zhu, Taesung Park, Phillip Isola, and Alexei~A Efros.
\newblock Unpaired image-to-image translation using cycle-consistent
  adversarial networks.
\newblock In {\em IEEE International Conference on Computer Vision}, pages
  2223--2232, 2017.

\bibitem{zhu2018visual}
Jun-Yan Zhu, Zhoutong Zhang, Chengkai Zhang, Jiajun Wu, Antonio Torralba, Josh
  Tenenbaum, and Bill Freeman.
\newblock Visual object networks: Image generation with disentangled 3d
  representations.
\newblock In {\em Advances in Neural Information Processing Systems},
  volume~31, pages 118--129, 2018.

\end{thebibliography}
}

\clearpage

\appendix

\begin{strip}
\centering
\Large{\textbf{Supplementary Material}}
\end{strip}

\renewcommand{\thesection}{\Alph{section}}
\renewcommand{\thefigure}{\Roman{figure}}
\renewcommand{\thetable}{\Roman{table}}
\renewcommand{\theequation}{\Roman{equation}}
\setcounter{figure}{0}
\setcounter{equation}{0}

\begin{strip}
	\centering
	\includegraphics[width=1\linewidth]{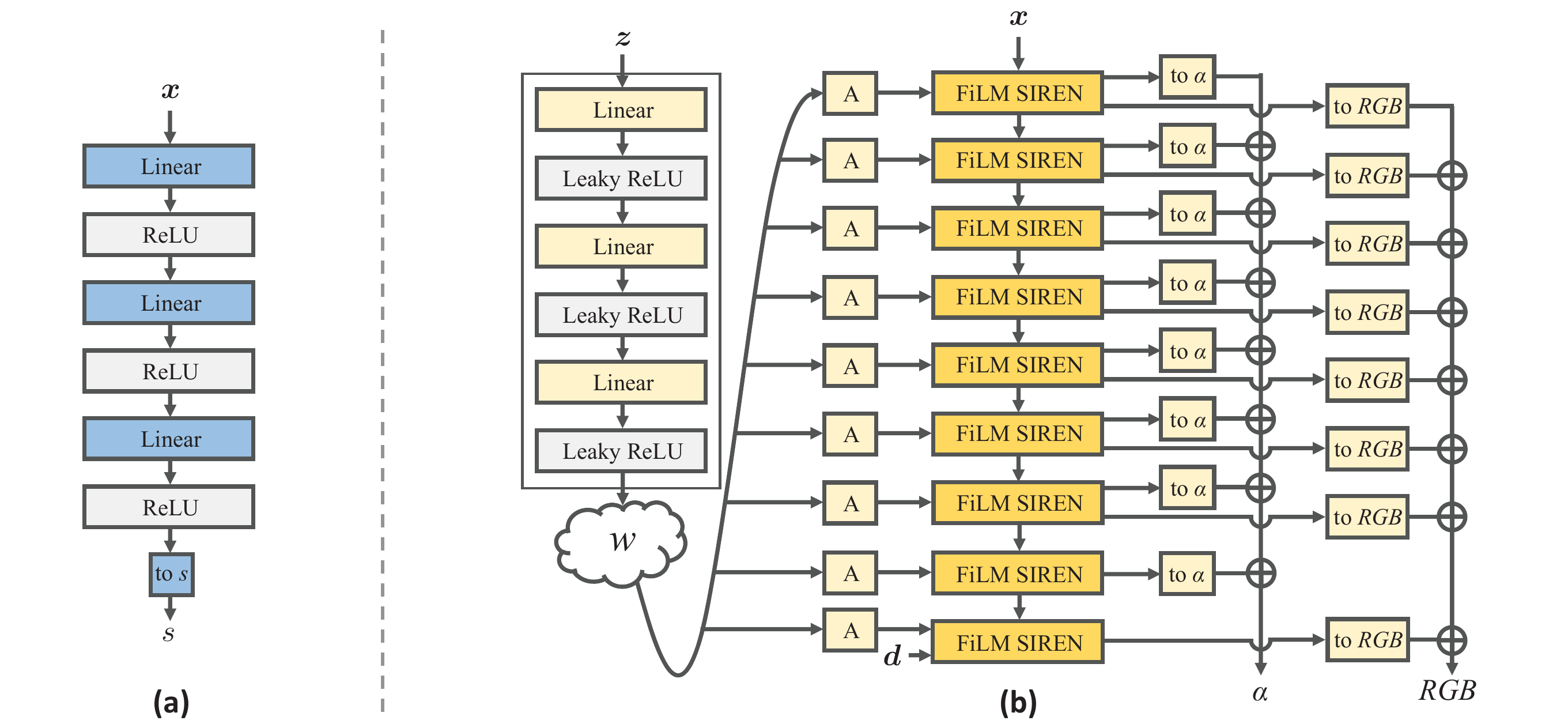}
	\vspace{-15pt}
	\captionsetup{type=figure,font=small}
	\caption{
		Detailed network structures of (a) the manifold predictor $\mathcal{M}$ and (b) the radiance generator $\Phi$.\label{fig:backbone}
	}
	\vspace{0pt}
\end{strip}

\section{More Implementation Details} \label{sec:implement}

\subsection{Data Preparation}

\paragraph{FFHQ~\cite{karras2019style}.} We align the face images in FFHQ using 5 facial landmarks to centralize the faces and normalize their scales.
Specifically, we first detected 5 facial landmarks of the images using an off-the-shelf landmark detector~\cite{bulat2017far}. Then we follow \cite{deng2019accurate} to resize and crop the images by solving a least square problem between the detected keypoints and corresponding 3D keypoints derived from a 3D face model~\cite{paysan20093d}. For pose distribution estimation, the face reconstruction method of \cite{deng2019accurate} is applied to extract the face poses for all the training images. Gaussian distributions are then fitted on the extracted poses, which are defined by the yaw and pitch angles (standard deviation $0.3$ radians and $0.15$ radians, respectively). During GAN training, we sample camera pose from the distributions and generate images accordingly. The extracted poses also serve as the pseudo labels for the pose regularization term defined in Eq.~(\ref{eq:pose}) of the main paper.  

\vspace{-9pt}
\paragraph{Cats~\cite{zhang2008cat}.} For the cat images, we follow a similar procedure to align and resize the images using landmarks provided by the dataset~\cite{zhang2008cat}. We also estimate the camera pose by solving the least square problem between the provided 2D landmarks and a set of manually-selected 3D landmarks on a 3D cat mesh. We found the pose distribution is very close to face images in FFHQ, and thus we simply use the same Gaussian to sample poses during training.

\vspace{-9pt}
\paragraph{CARLA~\cite{dosovitskiy2017carla,schwarz2020graf}.} We directly resize the car images rendered by \cite{schwarz2020graf} to $128^2$ resolution without any alignment. Following \cite{schwarz2020graf,chan2021pi}, we uniformly sample camera pose from the upper hemisphere during training.

\subsection{Network Structure}

\paragraph{Manifold predictor $\mathcal{M}$.} Figure~\ref{fig:backbone} (a) shows the structure of the manifold predictor, which is an MLP with three hidden layers and an output layer. We set the channel dimension of the hidden layers to 128, 64, and 256 for FFHQ, Cats, and CARLA, respectively. These channel dimensions are empirically chosen without careful tuning.

\vspace{-9pt}
\paragraph{Radiance generator $\Phi$.} Figure~\ref{fig:backbone} (b) shows the detailed structure of the radiance generator, which consists of a mapping network and a synthesis network. The mapping network is an MLP with three hidden layers of dimension 256. The synthesis network consists of $8$ FiLM SIREN blocks~\cite{chan2021pi} of dimension 256, and one FiLM SIREN block of dimension 259 which receives an extra view direction as input.

\subsection{More Training Details}
During training, we randomly sample latent code ${\bm z}$ from the normal distribution and camera pose ${\bm \theta}$ from the known or estimated distributions of the training datasets. 
We jointly learn the manifold predictor $\mathcal{M}$, the radiance generator $\Phi$, and the discriminator $D$ using the losses described in the main paper. Geometric initialization~\cite{atzmon2020sal} is applied for the weights of $\mathcal{M}$ to obtain sphere-like initial isosurfaces. 
For FFHQ and Cats, we set the sphere center to $(0,0,-1.5)$ for human face and cat centered in the $[-1,1]^3$ cube.
For CARLA, we set the center to $(0,0,0)$ to obtain hemispherical manifolds, as shown in Fig.~\ref{fig:manifold_vis} of the main paper. 
The $\{l_i\}$ are set to generate initial isosurfaces evenly positioned across the whole 3D volume.
In addition, for FFHQ and Cats, we set the farmost surface
to be a fixed plane to represent background. 
To calculate ray-surface intersections, we uniformly sample 64 points along each ray and calculate the intersections via Eq.~(\ref{eq:interpolate}) in the main paper.
The weights of the radiance generator $\Phi$ and the discriminator $D$ are initialized following  \cite{chan2021pi}.

To enable training at $256^2$ resolution, we use PyTorch's Automatic Mixed Precision (AMP) to reduce memory cost. We also use the mini-batch aggregation strategy similar to~\cite{chan2021pi} to ensure a relatively large batch size (16 for $256^2$ resolution and 32 for $128^2$ resolution) during training. We train GRAM for $120$K iterations, $80$K iterations, and $70$K iterations on FFHQ, Cats, and CARLA, respectively. Training took 3 to 7 days depending on the dataset and image resolution. 


\begin{figure}[t]
	\centering
	\includegraphics[width=1.0\columnwidth]{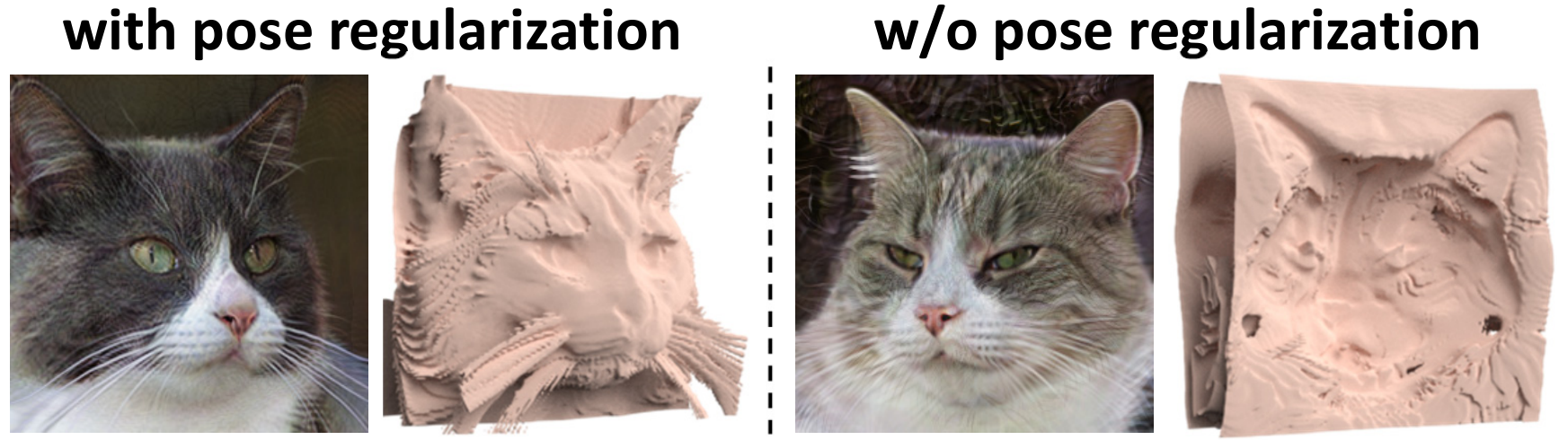}
 	\vspace{-13pt}
	\caption{Learned 3D geometry with and w/o pose regularization.}
	\label{fig:concave}
	\vspace{-3pt}
\end{figure}
\begin{figure}[t]
	\centering
	\includegraphics[width=1.0\columnwidth]{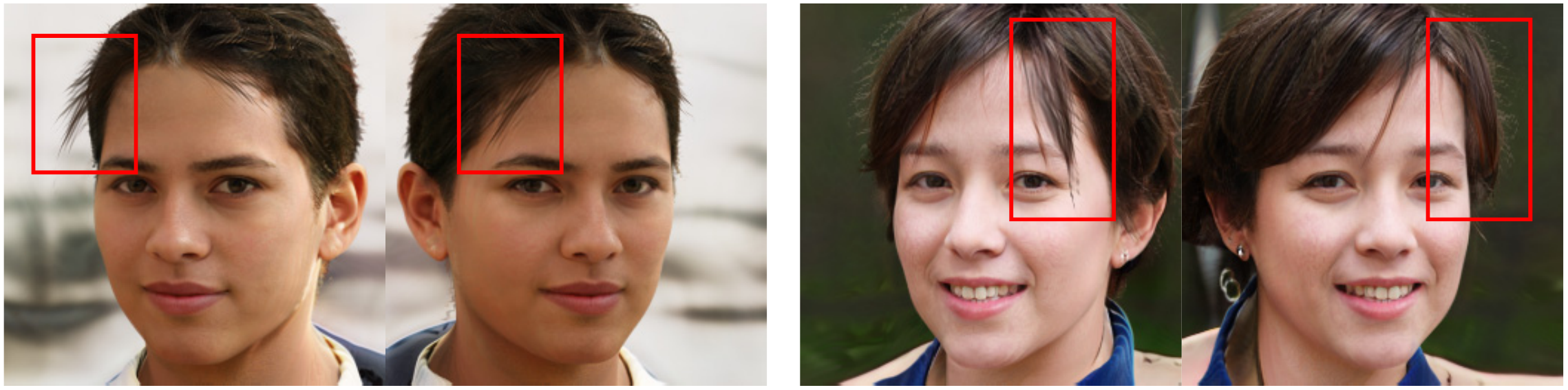}
 	\vspace{-13pt}
	\caption{Exaggerated parallax artifacts on generated subjects.}
	\label{fig:parallax}
	\vspace{-3pt}
\end{figure}

\section{More Results}

\subsection{Qualitative Results}
\begin{figure*}[t]
	\small
	\centering
	\includegraphics[width=0.92\textwidth]{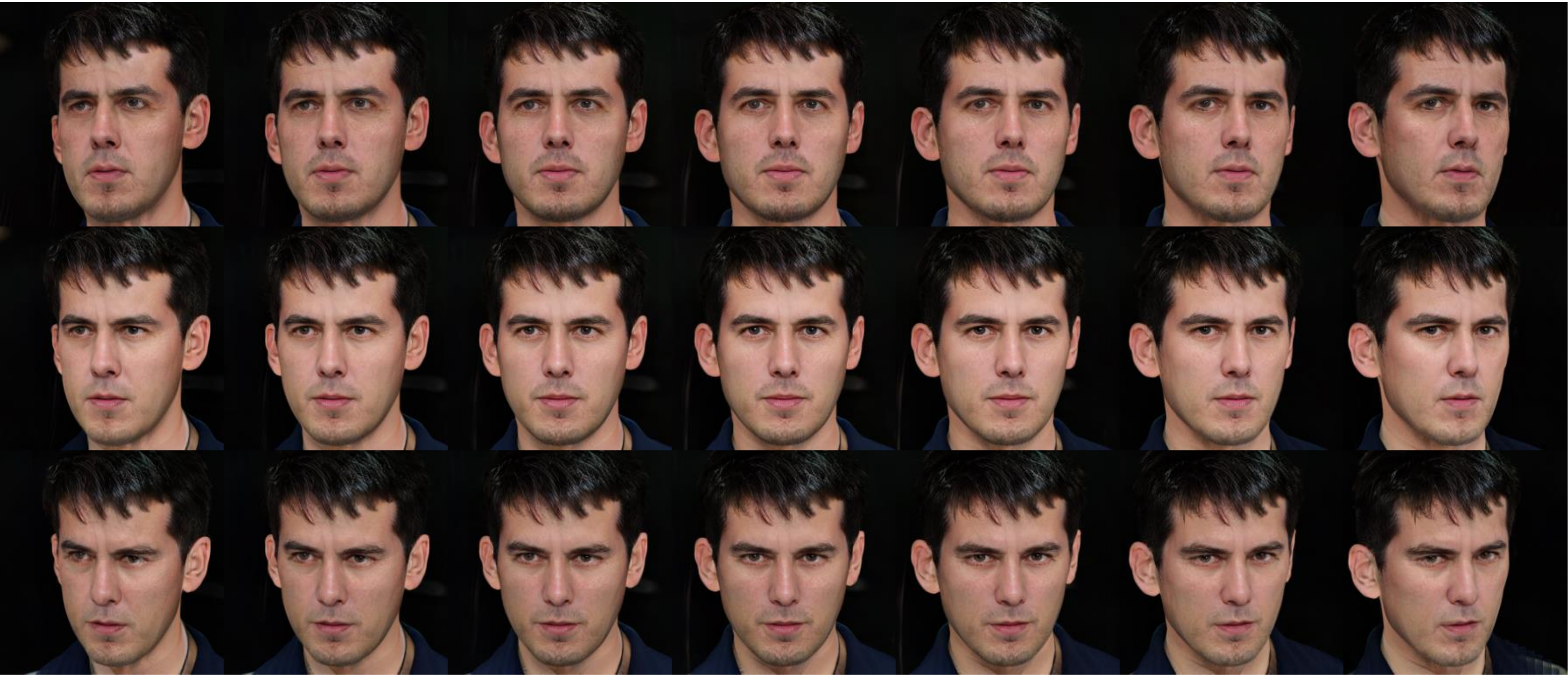}
	\includegraphics[width=0.92\textwidth]{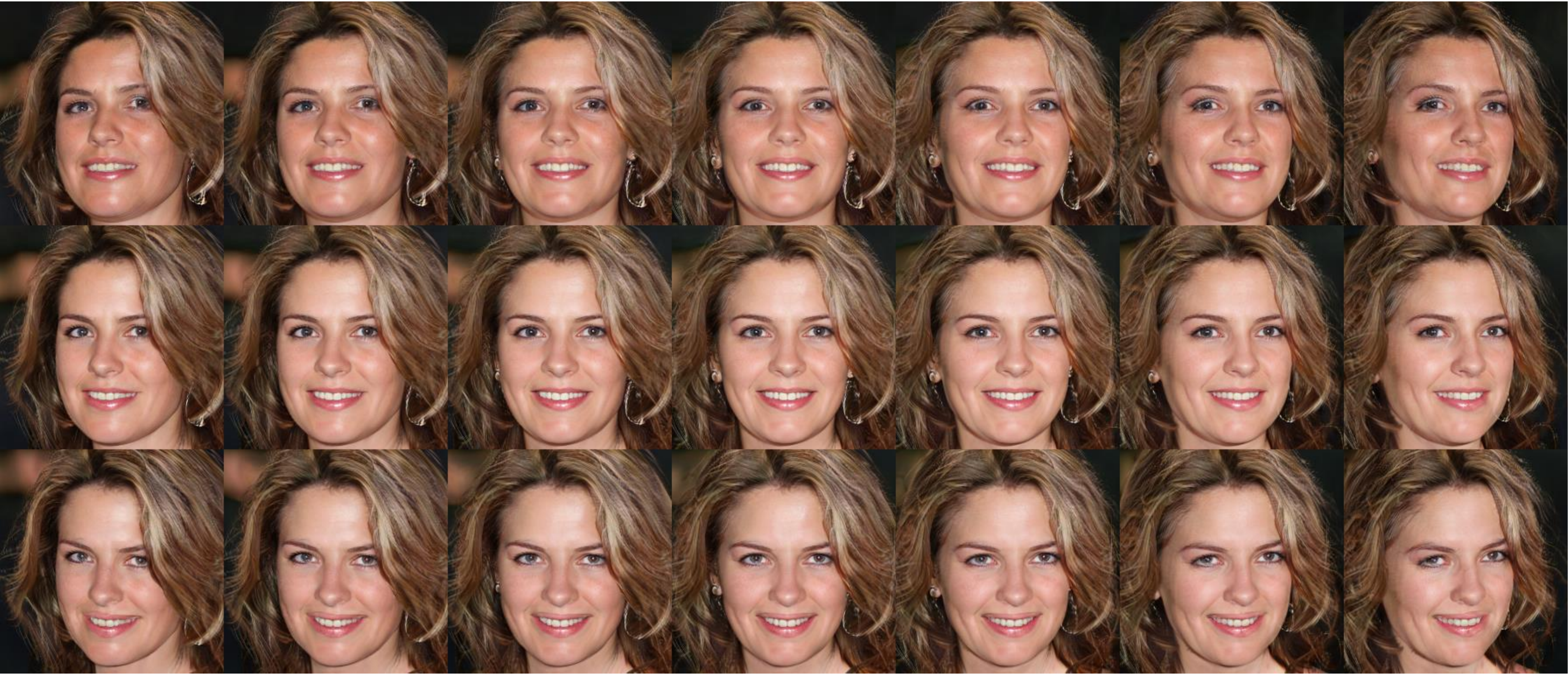}
	\includegraphics[width=0.92\textwidth]{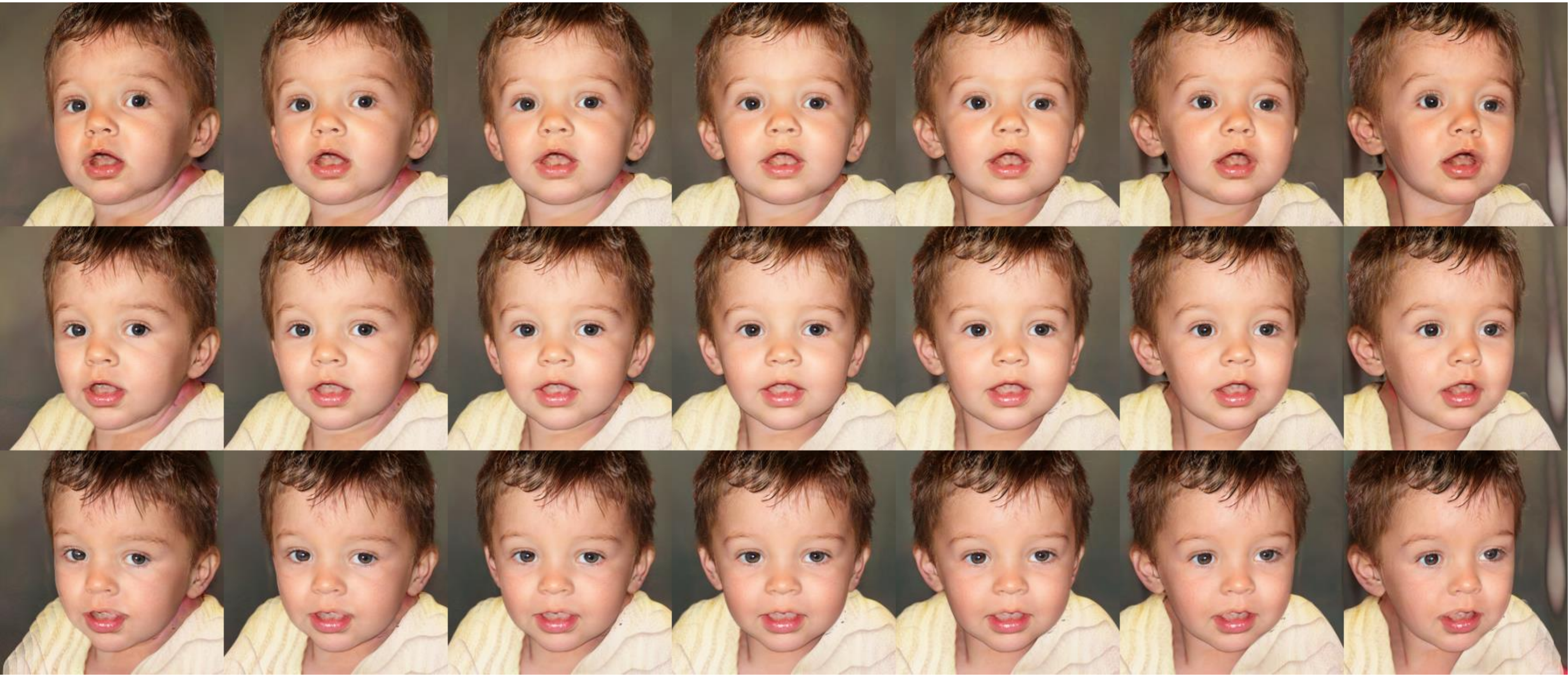}
 	\vspace{-5pt}
	\caption{Multiview generation results of GRAM on FFHQ.}
	\label{fig:multiview}
	\vspace{-3pt}
\end{figure*}

\begin{figure*}[t]
	\small
	\centering
	\includegraphics[width=0.92\textwidth]{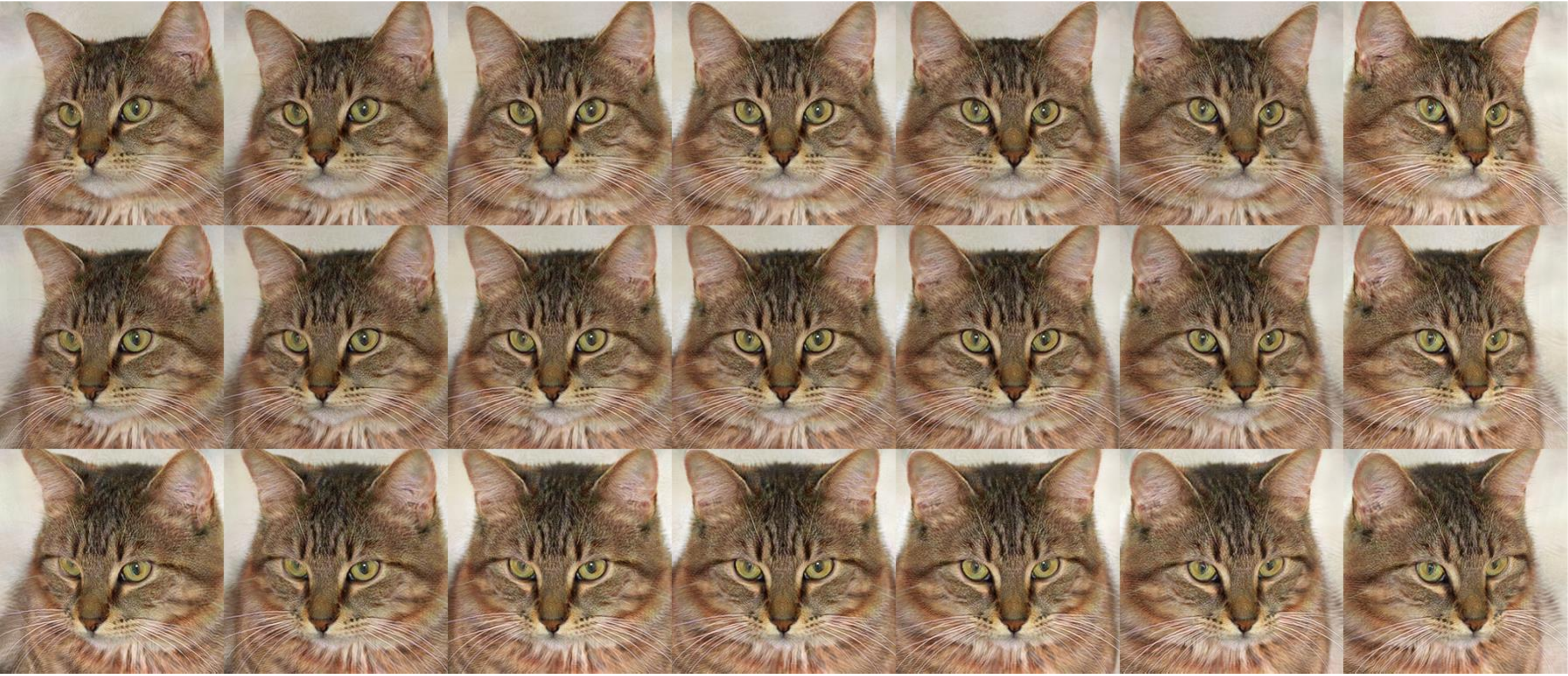}
	\includegraphics[width=0.92\textwidth]{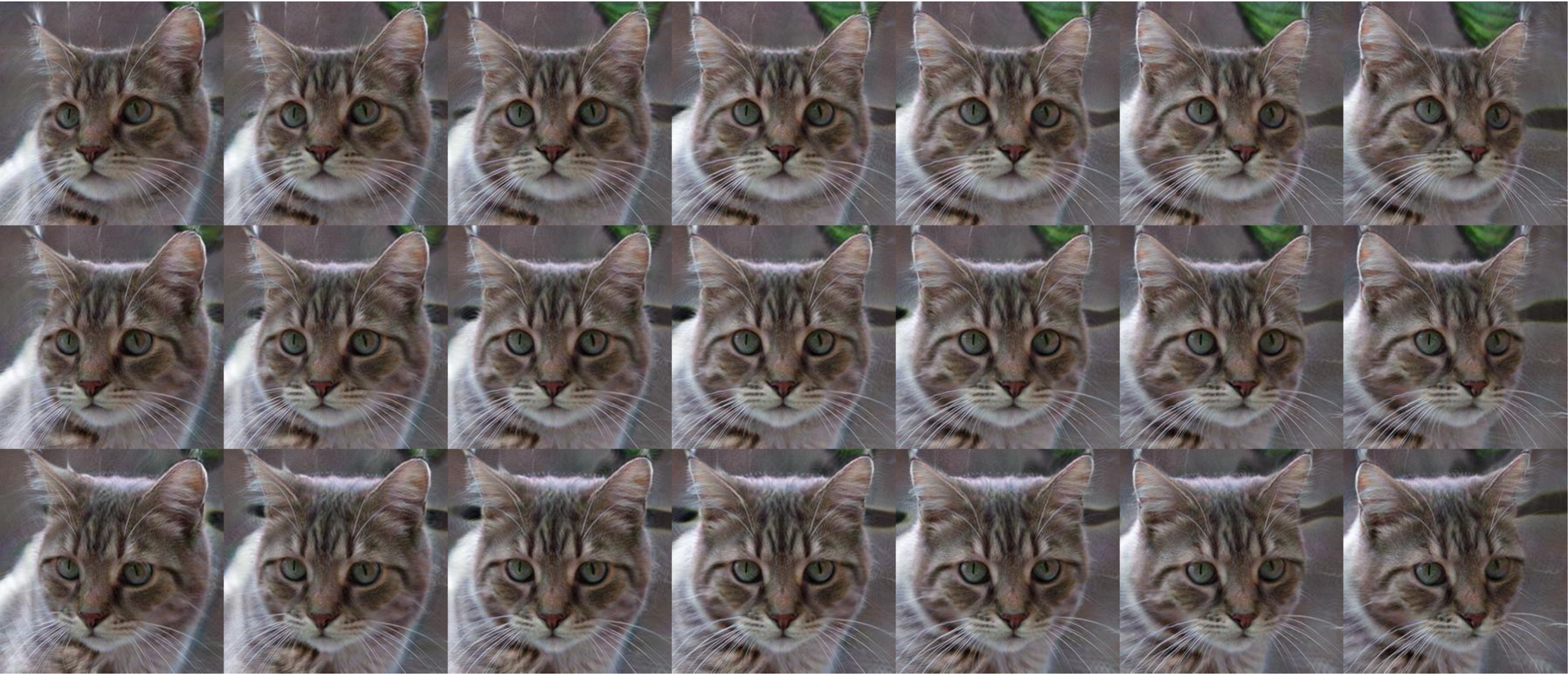}
	\includegraphics[width=0.92\textwidth]{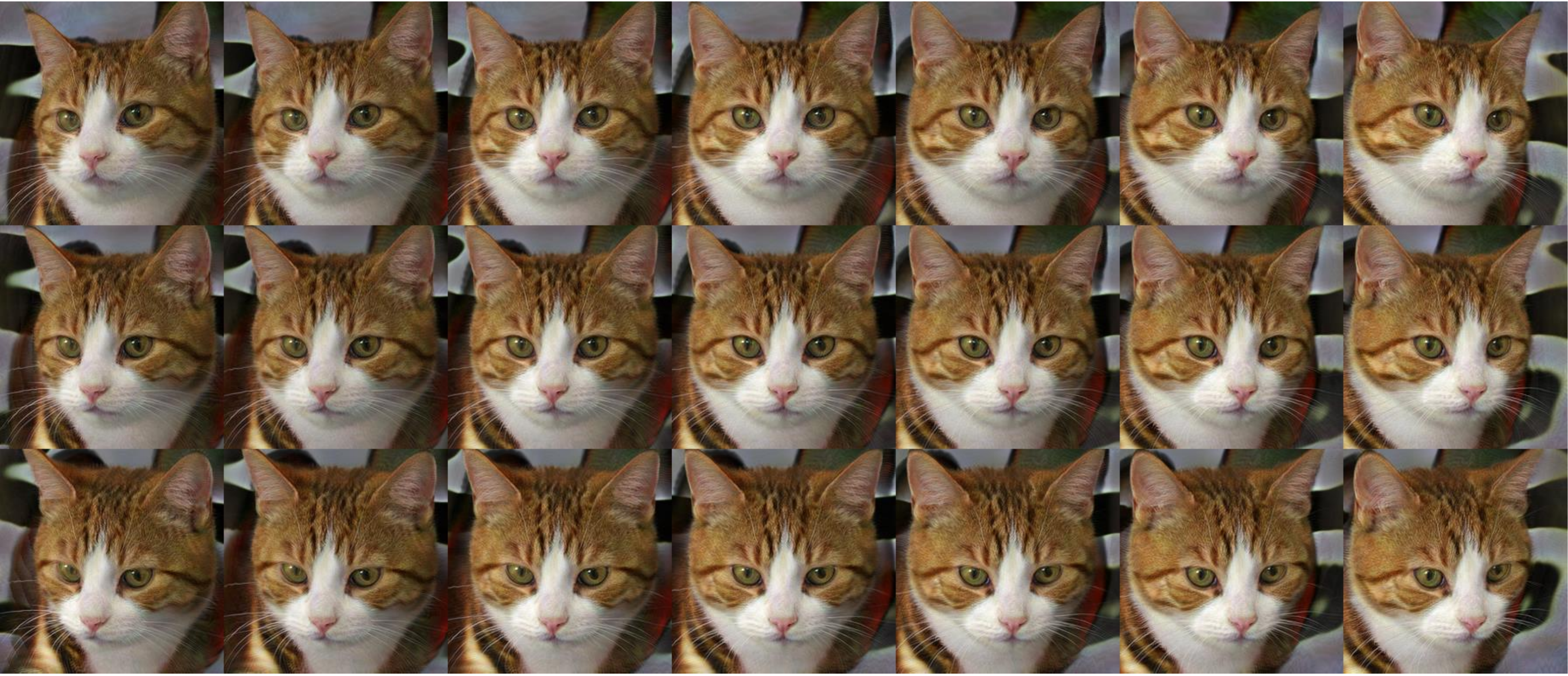}
 	\vspace{-5pt}
	\caption{Multiview generation results of GRAM on Cats.}
	\label{fig:multiview2}
	\vspace{-3pt}
\end{figure*}

\begin{figure*}[t]
	\small
	\centering
	\includegraphics[width=0.92\textwidth]{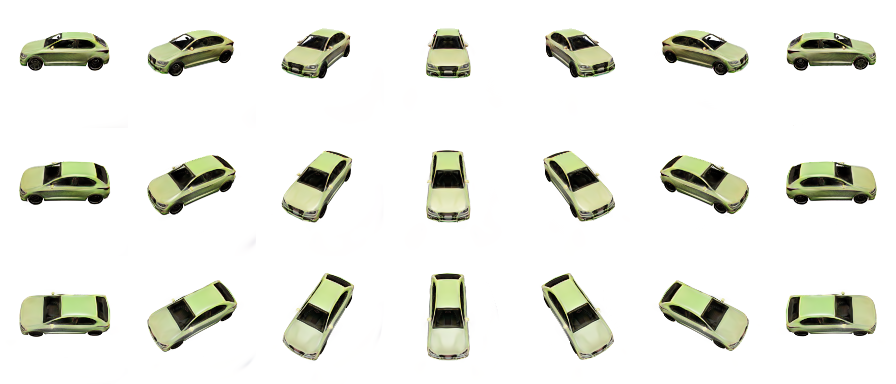}
	\includegraphics[width=0.92\textwidth]{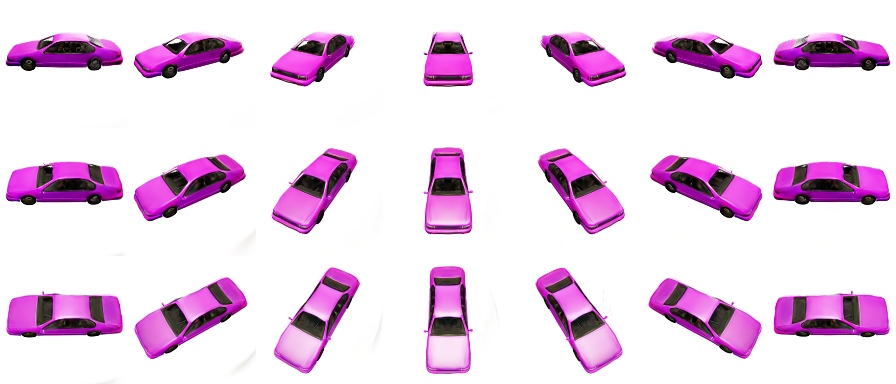}
	\includegraphics[width=0.92\textwidth]{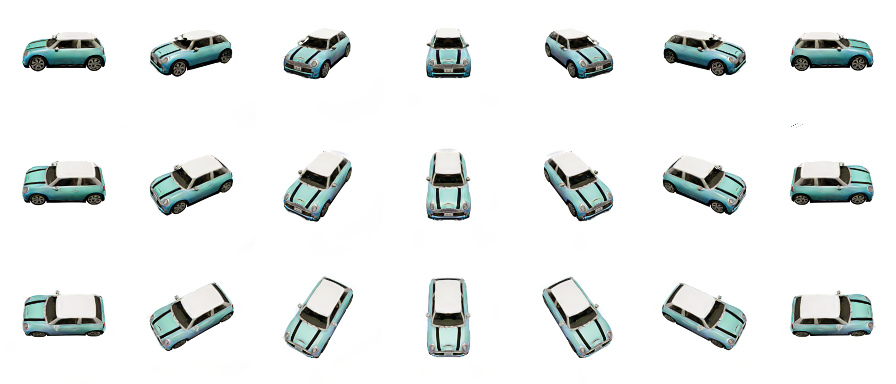}
 	\vspace{-5pt}
	\caption{Multiview generation results of GRAM on CARLA.}
	\label{fig:multiview3}
	\vspace{-3pt}
\end{figure*}

Figure~\ref{fig:multiview}, \ref{fig:multiview2}, and \ref{fig:multiview3} show more visual results of GRAM. Our method can generate realistic images with strong multiview consistency. Animation results can be found on the \href{https://yudeng.github.io/GRAM/}{\emph{project page}}.

\subsection{Comparisons}
\begin{figure*}[t]
	\small
	\centering
	\includegraphics[width=0.97\textwidth]{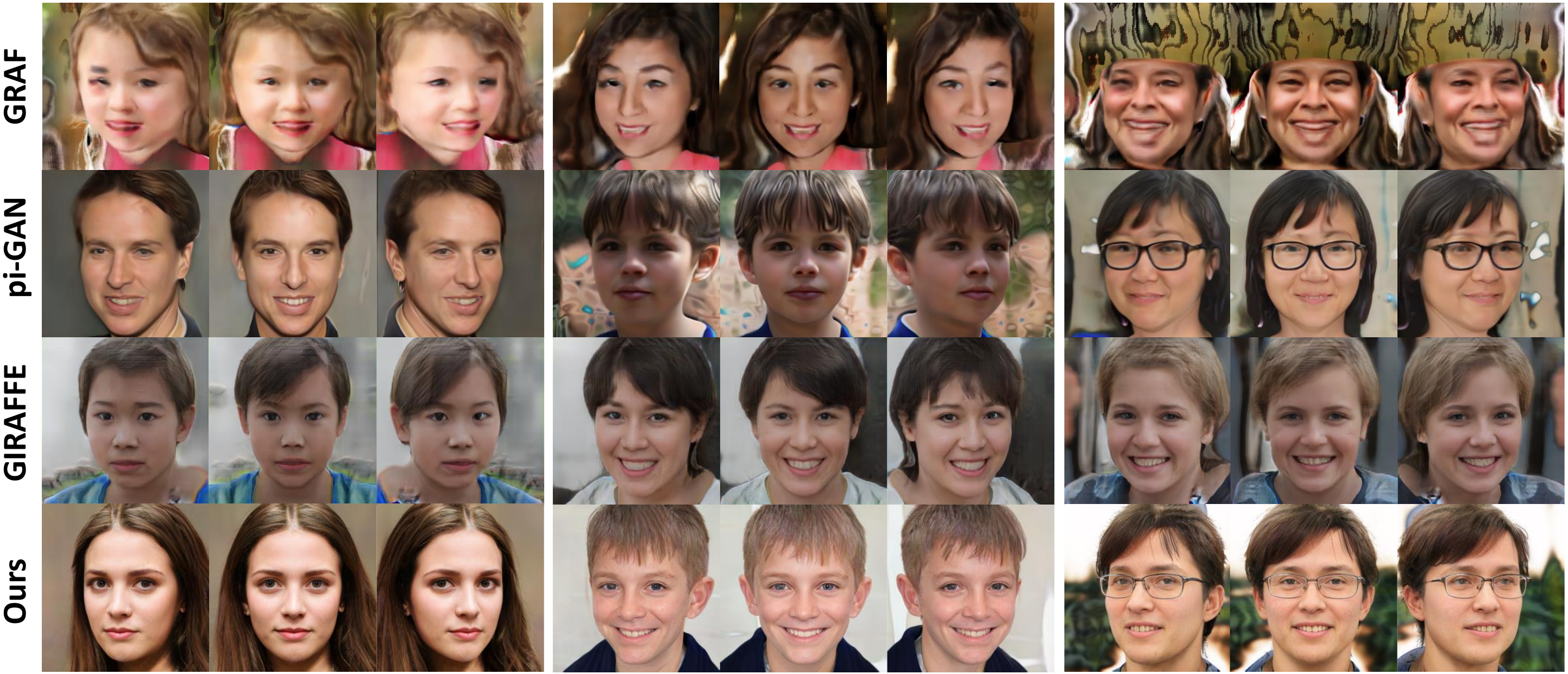}
	\includegraphics[width=0.97\textwidth]{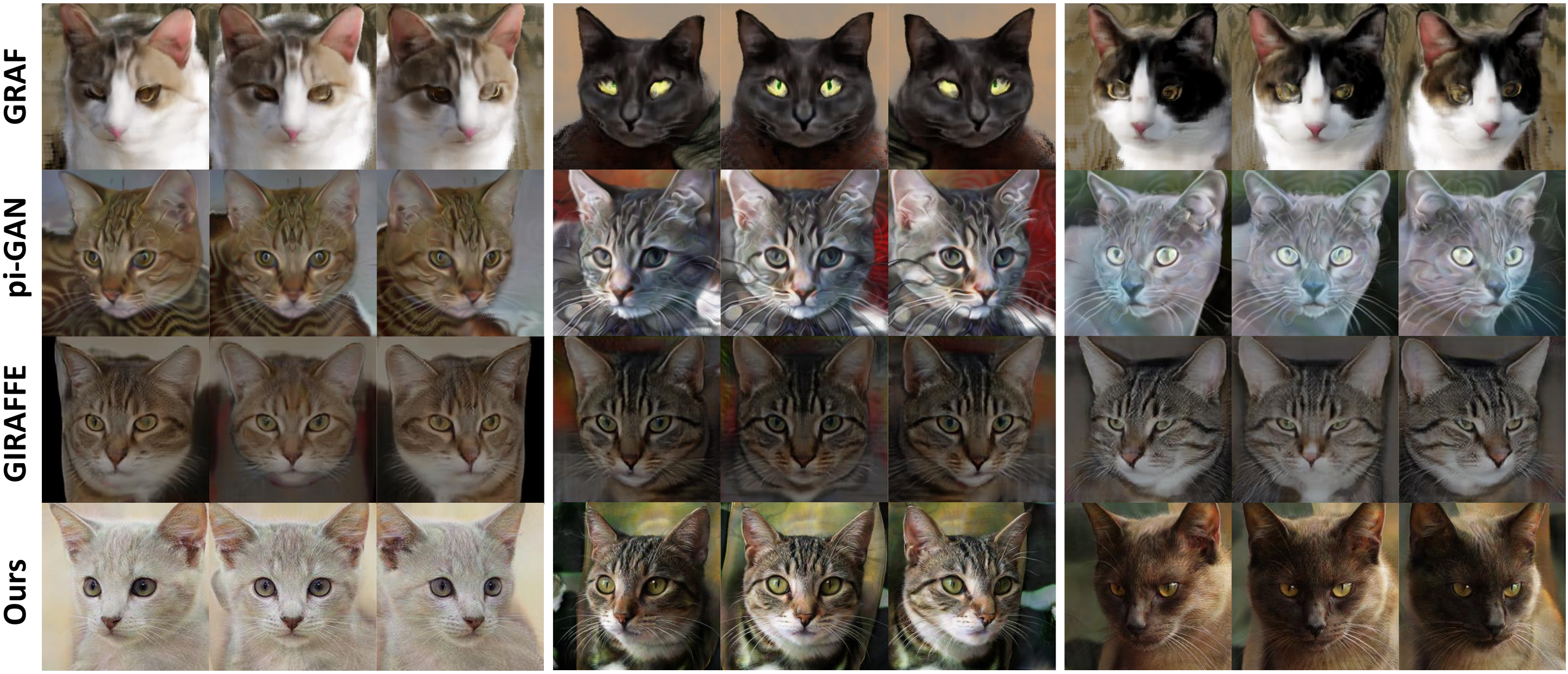}
	\includegraphics[width=0.97\textwidth]{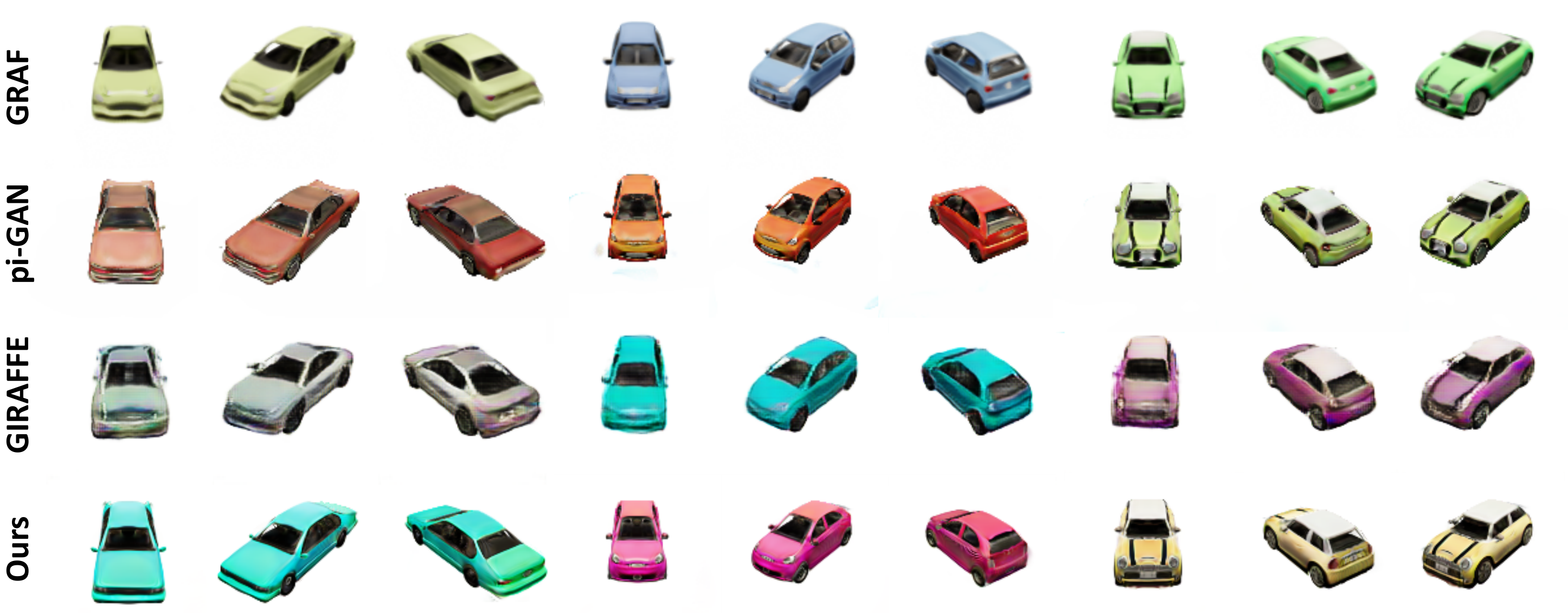}
	\caption{More qualitative comparisons with previous 3D-aware image generation methods on three datasets.}
	\label{fig:compare_more}
	\vspace{-3pt}
\end{figure*}

\paragraph{More comparisons with previous methods.}
Figure~\ref{fig:compare_more} shows more visual comparisons between GRAM and the previous 3D-aware image generation methods~\cite{schwarz2020graf,chan2021pi,niemeyer2021giraffe}. Our method achieves the best result in terms of image quality and 3D consistency. Animations can be found on the \href{https://yudeng.github.io/GRAM/}{\emph{project page}}.

\begin{figure*}[t]
	\small
	\centering
	\includegraphics[width=0.78\textwidth]{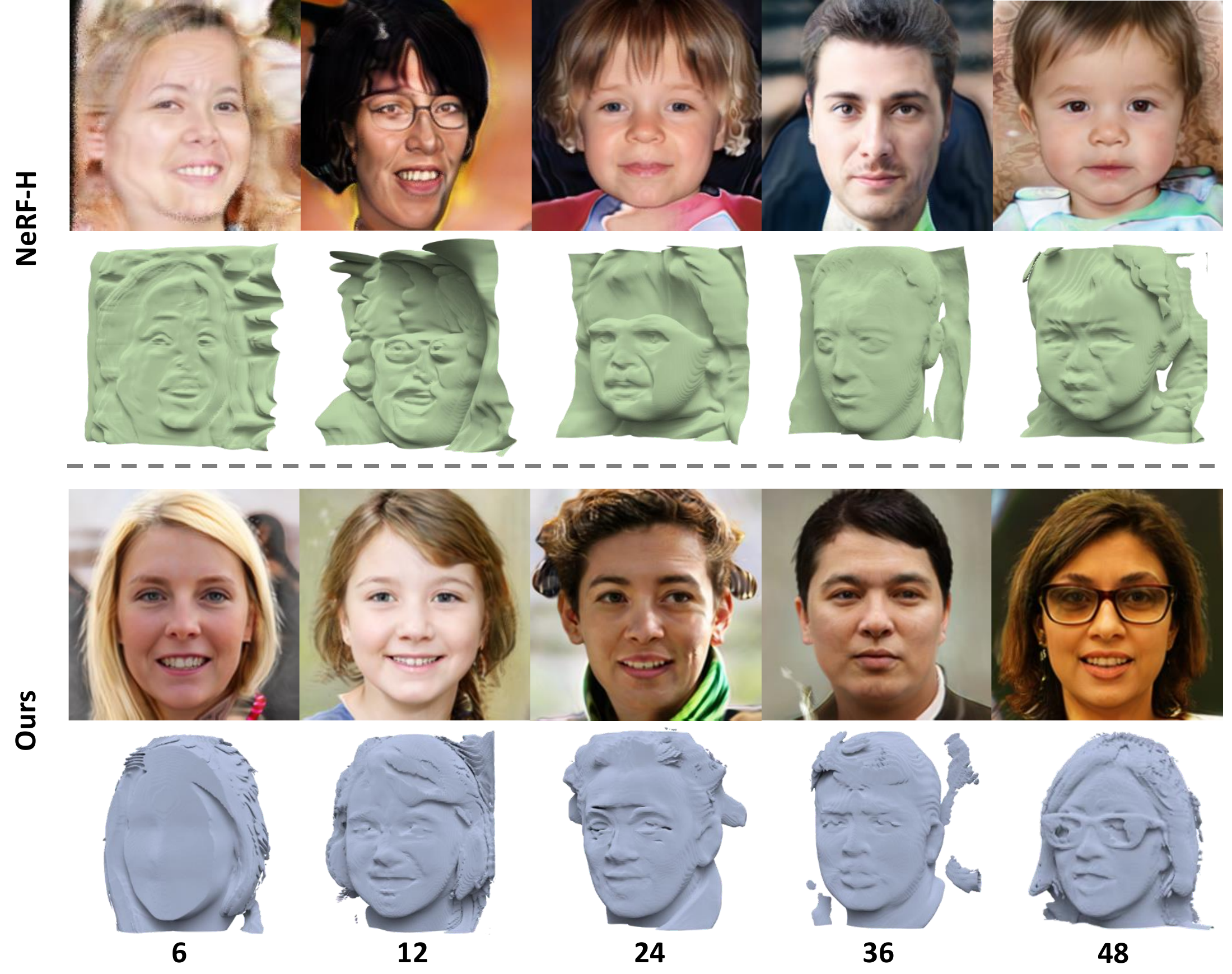}
	\vspace{-5pt}
	\caption{Comparison between our manifold sampling and NeRF-H~\cite{mildenhall2020nerf,chan2021pi} sampling strategy.}
	\label{fig:number}
	\vspace{0pt}
\end{figure*}

\vspace{-8pt}
\paragraph{More comparisons with NeRF-H sampling.}
Figure~\ref{fig:number} shows the visual comparisons between our manifold sampling strategy and the original NeRF-H~\cite{mildenhall2020nerf,chan2021pi} sampling strategy. Our method achieves better visual quality with finer details. More importantly, NeRF-H fails to learn reasonable 3D structures of the generated instances with a number of sampling points fewer than 12. It still produces undesired artifacts (\eg, the concave forehead geometry which creates hollow-face
illusion), even trained with 48 sampling points. In contrast, our method can learn reasonable 3D geometry with as few as 6 points (surfaces). We hardly observe the concave forehead issue for the generated instances in our cases.

\subsection{Failure Cases}\label{sec:failure}
\paragraph{Concave geometry.} We empirically found that for cats, dropping pose regularization sometimes led to unstable training and
yielded wrong pose and geometry (which is known as the “hollow-face illusion”; see Fig.~\ref{fig:concave}). Training on faces and cars were quite
stable no matter pose regularizations were used or not.

\paragraph{Exaggerated parallax artifacts.} When varying camera poses, some contents (\eg hair fringes) on certain generated subjects could float away from their expected positions, as shown in Fig.~\ref{fig:parallax}. This is due to that the fixed and limited number of surface manifolds across the whole category cannot provide accurate depth for all structures on every single subject. The problem could be alleviated when using instance-specific surfaces, which we will explore in future works.

\subsection{Camera Zoom}
\begin{figure*}[t]
	\small
	\centering
	\includegraphics[width=0.76\textwidth]{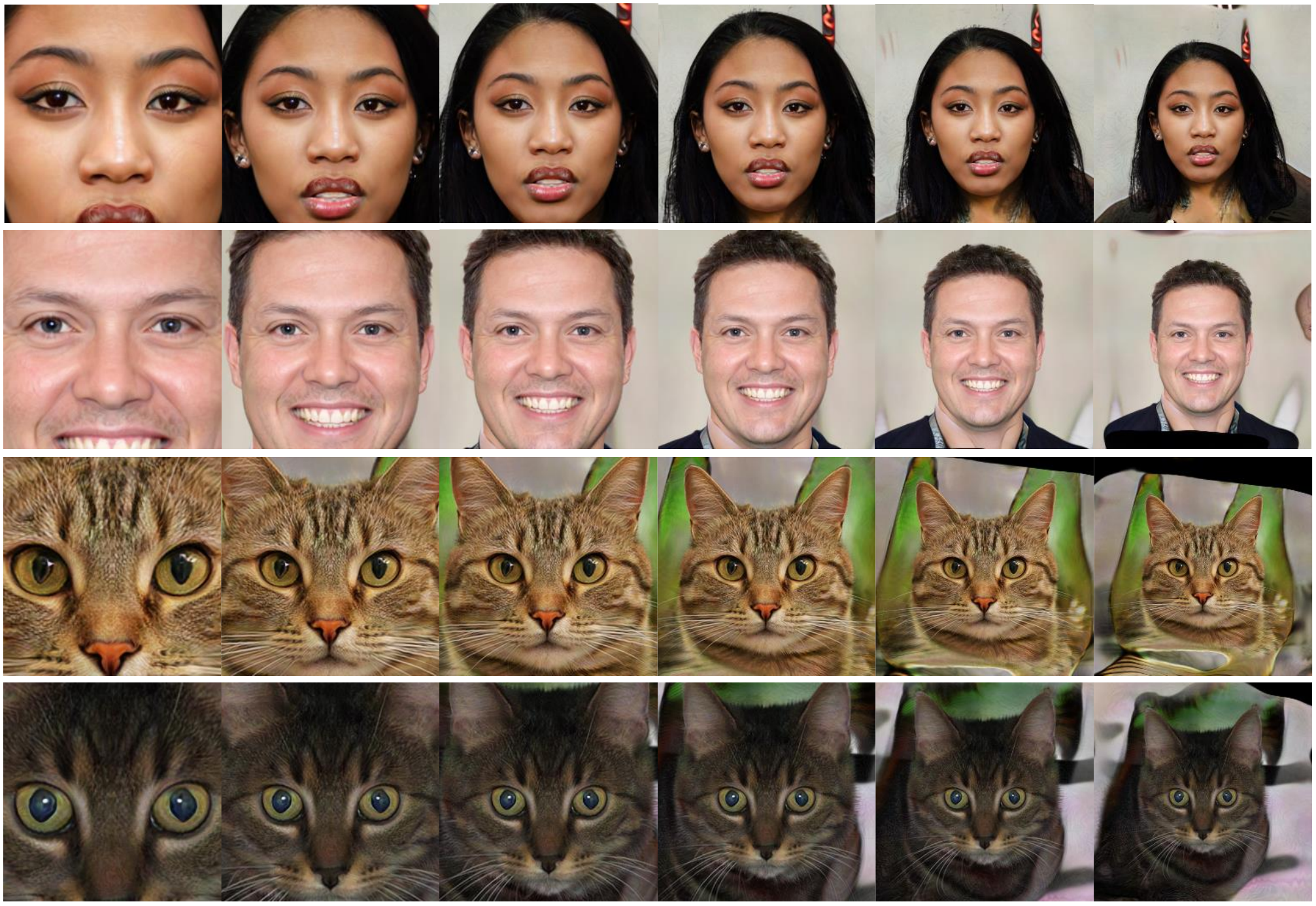}
 	\vspace{-8pt}
	\caption{Generation results under camera zoom-in and zoom-out.}
	\label{fig:zoom}
	\vspace{-3pt}
\end{figure*}

As shown in Fig.~\ref{fig:zoom}, GRAM can generate reasonable results with camera zoom-in and zoom-out effects. Animations can be found on the \href{https://yudeng.github.io/GRAM/}{\emph{project page}}.

\subsection{Latent Space Interpolation}
\begin{figure*}[t]
	\small
	\centering
	\includegraphics[width=1.0\textwidth]{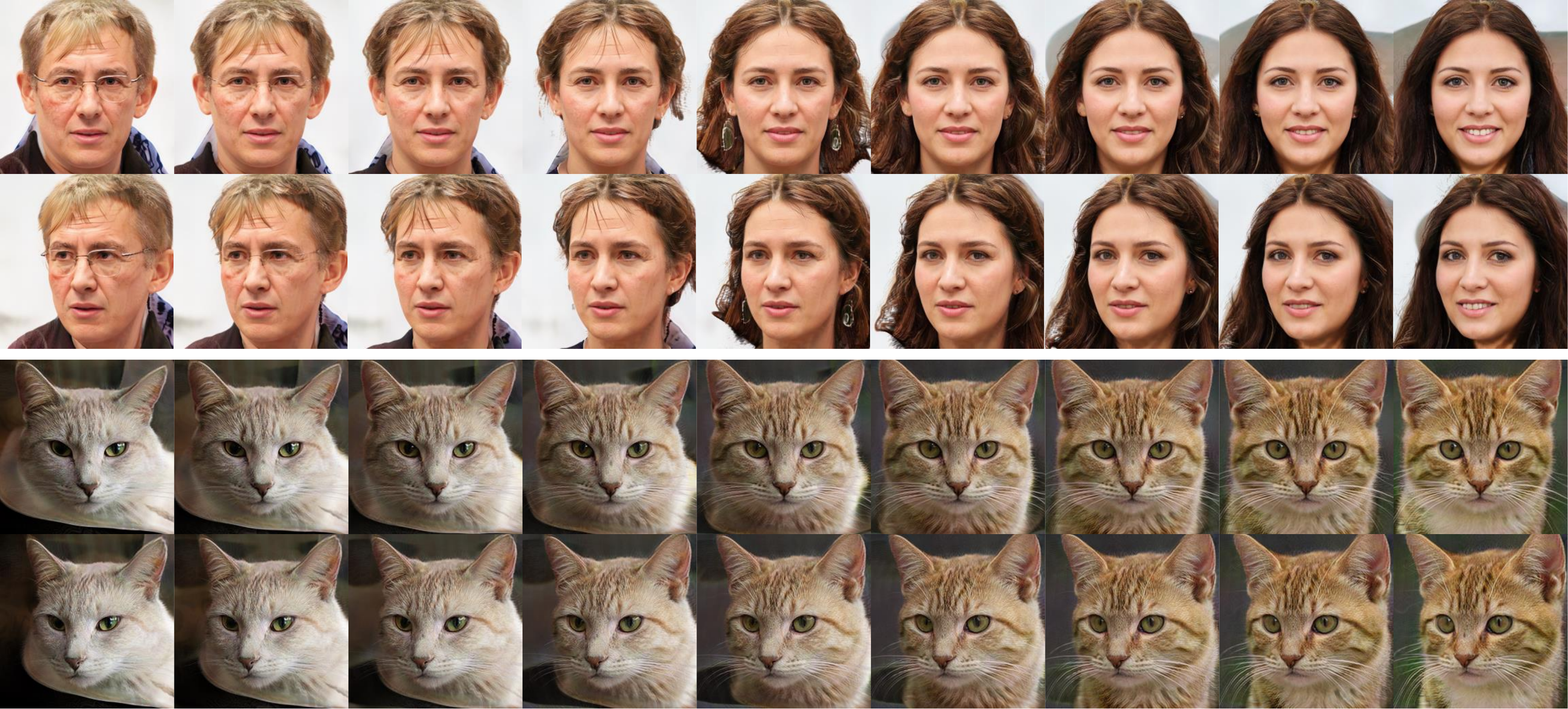}
	\vspace{-17pt}
	\caption{Latent space interpolation results.}
	\label{fig:interpolate}
	\vspace{20pt}
\end{figure*}

We show the results of latent code interpolation in Fig.~\ref{fig:interpolate}. The continuous semantic changes between adjacent images demonstrate the reasonable latent space learned by GRAM.

\subsection{Style Mixing}
\begin{figure*}[t]
	\small
	\centering
	\includegraphics[width=1.0\textwidth]{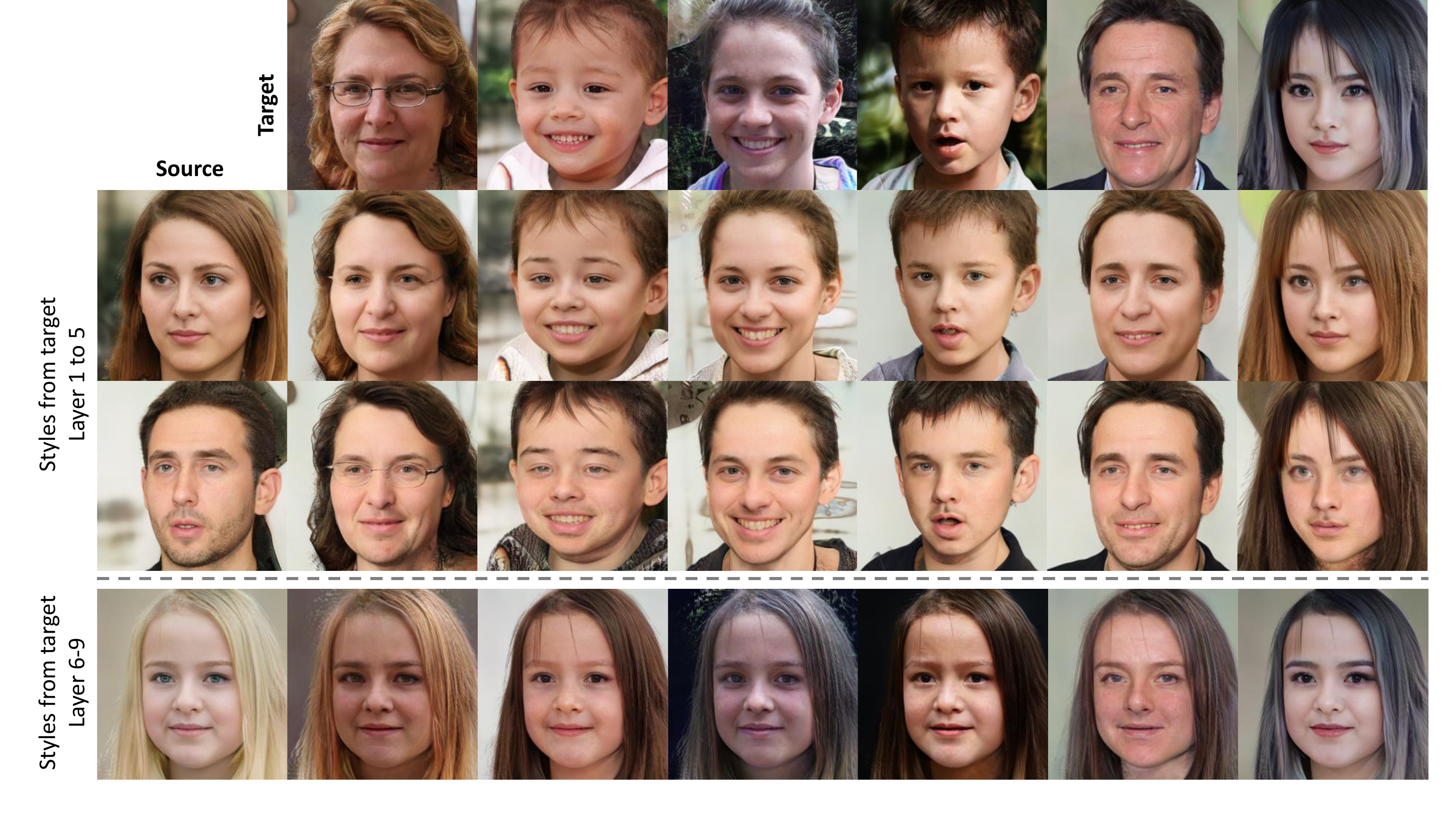}
	\vspace{-17pt}
	\caption{Style mixing between different generated subjects. Note that our method is not trained with the style mixing strategy.}
	\label{fig:style}
\end{figure*}

Figure~\ref{fig:style} shows the style mixing results between source subjects and target subjects. Similar to~\cite{karras2019style,karras2020analyzing}, styles in shallower layers (layer 1 to 5) of GRAM mainly control geometry, while styles in deeper layers (layer 6 to 9) control appearance. Note that our method is not trained with the style mixing strategy.


\end{document}